\pdfoutput=1

\documentclass[11pt]{article}

\usepackage{acl}

\usepackage{times}
\usepackage{latexsym}

\usepackage[T1]{fontenc}

\usepackage[utf8]{inputenc}

\usepackage{microtype}
\usepackage{inconsolata}

\usepackage{graphicx}
\usepackage{tabularx}
\usepackage{adjustbox}
\usepackage{booktabs}
\usepackage{multirow}
\usepackage{amsmath}
\usepackage{cuted}
\usepackage{comment}
\usepackage{lipsum}
\usepackage{colortbl}
\usepackage{xcolor}
\usepackage{enumitem}
\usepackage{bbding}
\usepackage{subcaption}

\title{Tuning LLMs with Contrastive Alignment Instructions for Machine Translation in Unseen, Low-resource Languages}

\author{Zhuoyuan Mao\thanks{\ \ Currently at Sony Group Corporation. Work done during Apple internship.} \and Yen Yu \\
   Apple \\
  \texttt{kevinmzy@gmail.com, yen\_yu@apple.com}}

\begin{document}
\maketitle
\begin{abstract}
This article introduces contrastive alignment instructions (\textbf{AlignInstruct}) to address two challenges in machine translation (MT) on large language models (LLMs). One is the expansion of supported languages to previously unseen ones. The second relates to the lack of data in low-resource languages. Model fine-tuning through MT instructions (\textbf{MTInstruct}) is a straightforward approach to the first challenge. However, MTInstruct is limited by weak cross-lingual signals inherent in the second challenge. AlignInstruct emphasizes cross-lingual supervision via a cross-lingual discriminator built using statistical word alignments. Our results based on fine-tuning the BLOOMZ models (1b1, 3b, and 7b1) in up to 24 unseen languages showed that: (1) LLMs can effectively translate unseen languages using MTInstruct; (2) AlignInstruct led to consistent improvements in translation quality across 48 translation directions involving English; (3) Discriminator-based instructions outperformed their generative counterparts as cross-lingual instructions; (4) AlignInstruct improved performance in 30 zero-shot directions.

\end{abstract}

\section{Introduction}

Large language models (LLMs)~\cite{DBLP:conf/nips/BrownMRSKDNSSAA20,DBLP:journals/corr/abs-2204-02311,DBLP:journals/corr/abs-2211-05100,DBLP:journals/corr/abs-2302-13971,muennighoff-etal-2023-crosslingual,DBLP:journals/corr/abs-2303-08774,DBLP:journals/corr/abs-2305-10403,DBLP:journals/corr/abs-2307-09288} achieved good performance for a wide range of NLP tasks for prevalent languages. However, insufficient coverage for low-resource languages remains to be one significant limitation. Low-resource languages are either not present, or orders of magnitude smaller in size than dominant languages in the pre-training dataset. This limitation is in part due to the prohibitive cost incurred by curating good quality and adequately sized datasets for pre-training. Incrementally adapting existing multilingual LLMs to incorporate an unseen, low-resource language thus becomes a cost-effective priority to address this limitation. Previous study~\cite{DBLP:conf/sepln/RosaF22,DBLP:journals/corr/abs-2202-03371,yong-etal-2023-bloom} explored extending language support using either continual pre-training~\cite{neubig-hu-2018-rapid,artetxe-etal-2020-cross,muller-etal-2021-unseen,ebrahimi-kann-2021-adapt}, or parameter efficient fine-tuning (PEFT) methods~\cite{pfeiffer-etal-2020-mad,DBLP:conf/iclr/HuSWALWWC22,DBLP:conf/nips/LiuTMMHBR22} on monolingual tasks. Extending language support for cross-lingual tasks remains underexplored due to the challenge of incrementally inducing cross-lingual understanding and generation abilities in LLMs~\cite{yong-etal-2023-bloom}.

\begin{figure}[t]
    \centering
    \includegraphics[width=\linewidth]{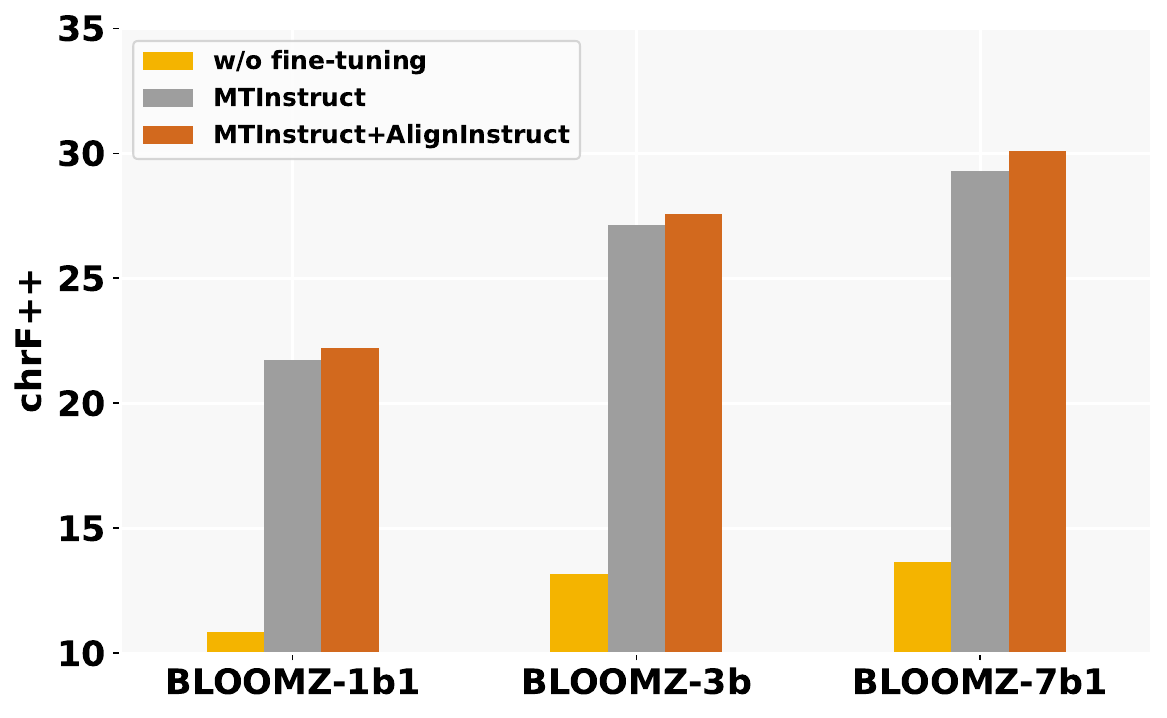}
    \caption{\textbf{Average chrF++ scores of BLOOMZ models across 24 unseen languages}, comparing settings of without fine-tuning, fine-tuning with MTInstruct, and fine-tuning that combines MTInstruct and AlignInstruct.}
    \label{fig:intro}
\end{figure}

This study focused on machine translation (MT) to highlight the cross-lingual LLM adaptation challenge. The challenge lies in enabling translation for low-resource languages that often lack robust cross-lingual signals. We first explored the efficacy of fine-tuning LLMs with MT instructions (MTInstruct) in unseen, low-resource languages. MTInstruct is a method previously shown to bolster the translation proficiency of LLMs for supported languages~\cite{DBLP:journals/corr/abs-2305-15083}. Subsequently, given that cross-lingual alignments are suboptimal in LLMs as a result of data scarcity of low-resource languages, we proposed contrastive alignment instructions (AlignInstruct) to explicitly provide cross-lingual supervision during MT fine-tuning. AlignInstruct is a cross-lingual discriminator formulated using statistical word alignments. Our approach was inspired by prior studies~\cite{DBLP:journals/mt/LambertPMW12,ren-etal-2019-explicit,lin-etal-2020-pre,mao-etal-2022-contrastive}, which indicated the utility of word alignments in enhancing MT. In addition to AlignInstruct, we discussed two word-level cross-lingual instruction alternatives cast as generative tasks for comparison with AlignInstruct.




Our experiments fine-tuned the BLOOMZ models~\cite{muennighoff-etal-2023-crosslingual} of varying sizes (1b1, 3b, and 7b1) for 24 unseen, low-resource languages, and evaluated translation on OPUS-100~\cite{zhang-etal-2020-improving} and Flores-200~\cite{DBLP:journals/corr/abs-2207-04672}. We first showed that MTInstruct effectively induced the translation capabilities of LLMs for these languages. Building on the MTInstruct baseline, the multi-task learning combining AlignInstruct and MTInstruct resulted in stronger translation performance without the need for additional training corpora. The performance improved with larger BLOOMZ models, as illustrated in Fig.~\ref{fig:intro}, indicating that AlignInstruct is particularly beneficial for larger LLMs during MT fine-tuning. When compared with the generative variants of AlignInstruct, our results indicated that discriminative instructions better complemented MTInstruct. Furthermore, merging AlignInstruct with its generative counterparts did not further improve translation quality, underscoring the efficacy and sufficiency of AlignInstruct in leveraging word alignments for MT. 

In zero-shot translation evaluation on the OPUS benchmark, AlignInstruct exhibited improvements over the MTInstruct baseline in 30 zero-shot directions between non-English languages, when exclusively fine-tuned with three unseen languages (German, Dutch, and Russian). However, when incorporating supported languages (Arabic, French, and Chinese) the benefits of AlignInstruct were only evident in zero-shot translations where the target language was a supported language. In addition, to interpret the inherent modifications within the BLOOMZ models after applying MTInstruct or AlignInstruct, we conducted a visualization of the layer-wise cross-lingual alignment capabilities of the model representations. 




\section{Methodology}

\begin{figure*}[t]
    \centering
    \includegraphics[width=0.9\linewidth]{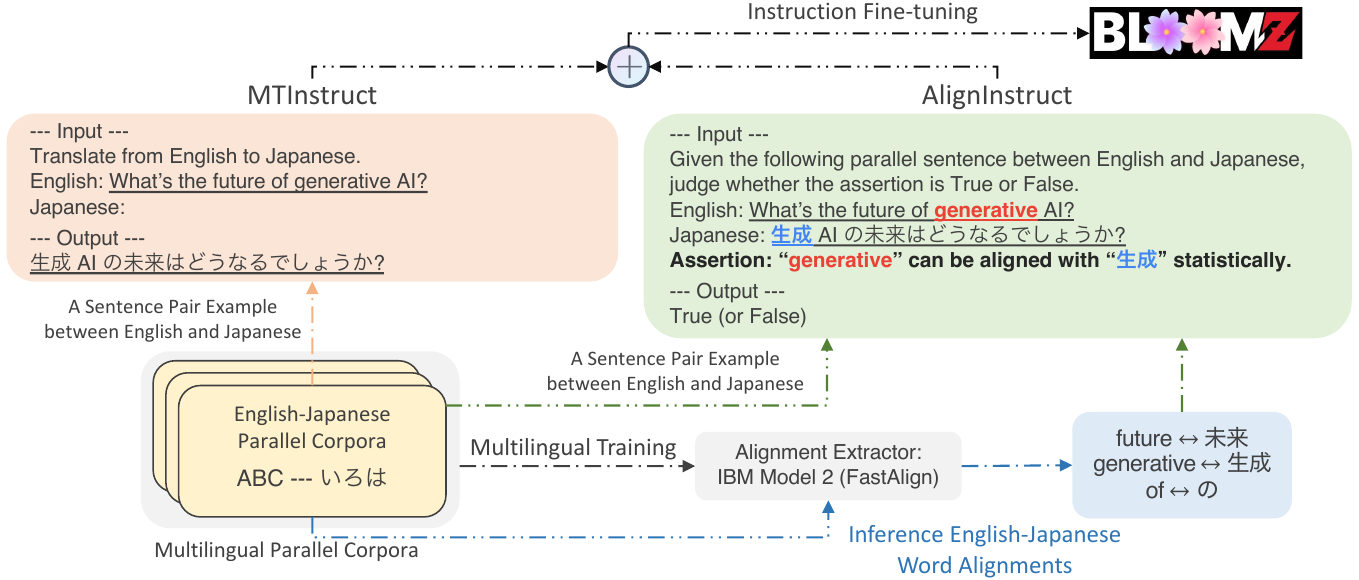}
    \caption{\textbf{Proposed instruction tuning methods combining MTInstruct (Sec.~\ref{sec:2.1}) and AlignInstruct (Sec.~\ref{sec:2.2}) for LLMs in MT tasks.} $\oplus$ denotes combining multiple instruction patters with a specific fine-tuning curriculum (Sec.~\ref{sec:3.2}). IBM Model 2 indicates word alignment model of statistical machine translation~\cite{brown-etal-1993-mathematics}.} 
    \label{fig:method}
\end{figure*}

This section presents MTInstruct as the baseline, and AlignInstruct. The MTInstruct baseline involved fine-tuning LLMs using MT instructions. AlignInstruct dealt with the lack of cross-lingual signals stemming from the limited parallel training data in low-resource languages. The expectation was enhanced cross-lingual supervision cast as a discriminative task without extra training corpora. Following this, we introduced two generative variants of AlignInstruct for comparison.\footnote{We also discussed monolingual instructions for MT fine-tuning in App.~\ref{app:mono}.}




\subsection{Baseline: MTInstruct}
\label{sec:2.1}
Instruction tuning~\cite{wang-etal-2022-super,mishra-etal-2022-cross,DBLP:journals/corr/abs-2210-11416,DBLP:conf/nips/Ouyang0JAWMZASR22,DBLP:conf/iclr/SanhWRBSACSRDBX22,DBLP:conf/iclr/WeiBZGYLDDL22} has been shown to generalize LLMs' ability to perform various downstream tasks, including MT~\cite{DBLP:journals/corr/abs-2305-15083}. 

Given a pair of the parallel sentences, $\left(\left(x_i\right)_1^N,\left(y_j\right)_1^M\right)$, where $(x_i)_1^N:=x_1 x_2 \ldots x_N$, $(y_j)_1^M:=y_1 y_2 \ldots y_M$. $x_i, y_j \in \mathcal{V}$ are members of the vocabulary $\mathcal{V}$ containing unique tokens that accommodate languages $X$ and $Y$.~\citet{DBLP:journals/corr/abs-2305-15083} showed that the following MT instructions (MTInstruct) can improve the translation ability in an LLM with a limited number of parallel sentences:

\begin{itemize}
    \setlength{\itemsep}{0pt}
    \setlength{\parskip}{0pt}
    \setlength{\topsep}{0pt}
    \item \textbf{Input:} ``Translate from $Y$ to $X$.
    
    $Y$: $y_1 y_2 \ldots y_M$.

    $X$: ''
    \item \textbf{Output}: ``$x_1 x_2 \ldots x_N$.''
\end{itemize}

Note that~\citet{DBLP:journals/corr/abs-2305-15083} demonstrated the utility of MTInstruct solely within the context of fine-tuning for languages acquired at pre-training phase. This study called for an assessment of MTInstruct on its efficacy for adapting to previously unsupported languages, denoted as $X$, accompanied by the parallel data in a supported language $Y$. 


\subsection{AlignInstruct}
\label{sec:2.2}
Word alignments have been demonstrated to enhance MT performance~\cite{DBLP:journals/mt/LambertPMW12,ren-etal-2019-explicit,lin-etal-2020-pre,mao-etal-2022-contrastive}, both in the fields of statistical machine translation (SMT)~\cite{brown-etal-1993-mathematics} and neural machine translation (NMT)~\cite{DBLP:conf/nips/SutskeverVL14,DBLP:journals/corr/BahdanauCB14}.~\citet{ren-etal-2019-explicit} and~\citet{mao-etal-2022-contrastive} reported the utility of SMT-derived contrastive word alignments in guiding encoder-decoder NMT model training. Built upon their findings, we introduced AlignInstruct for bolstering cross-lingual alignments in LLMs. We expected AlignInstruct to enhancing translation performance particularly for languages with no pre-training data and limited fine-tuning data.

As shown in Fig.~\ref{fig:method}, we employed FastAlign~\cite{dyer-etal-2013-simple} to extract statistical word alignments from parallel corpora. Our approach depended on a trained FastAlign model~\cite[IBM Model 2,][]{brown-etal-1993-mathematics} to ensure the quality of the extracted word pairs. These high-quality word alignment pairs were regarded as ``gold'' word pairs for constructing AlignInstruct instructions.\footnote{Note that these word pairs may not necessarily represent direct translations of each other; instead, they are word pairs identified based on their co-occurrence probability within the similar context. Refer to IBM model 2 in SMT.} Assuming one gold word pair $\left(x_k x_{k+1}, y_l y_{l+1} y_{l+2}\right)$ was provided for the sentence pair $\left(\left(x_i\right)_1^N,\left(y_j\right)_1^M\right)$, the AlignInstruct instruction reads:

\begin{itemize}
    \setlength{\itemsep}{0pt}
    \setlength{\parskip}{0pt}
    \setlength{\topsep}{0pt}
    \item \textbf{Input:} ``Given the following parallel sentence between $Y$ and $X$, judge whether the assertion is True or False.
    
    $Y$: $y_1 y_2 \ldots y_M$.

    $X$: $x_1 x_2 \ldots x_N$.

    Assertion: ``$y_l y_{l+1} y_{l+2}$'' can be aligned with ``$x_k x_{k+1}$'' statistically.''
    \item \textbf{Output}: ``True'' (or ``False'')
\end{itemize}

Instructions with the ``False'' output were constructed by uniformly swapping out part of the word pair to create misalignment. We anticipated that this treatment forced the model to learn to infer the output by recognizing true alignment-enriched instructions. This would require the model to encode word-level cross-lingual representation, a crucial characteristic for MT tasks.

\subsection{Generative Counterparts of AlignInstruct}
Previous studies~\cite{liang-etal-2022-contrastive,DBLP:journals/corr/abs-2305-06176} have suggested the importance of both discriminative and generative tasks in fine-tuning LLMs. We accordingly considered two generative variants of AlignInstruct. We then compared them with AlignInstruct to determine the most effective training task. As detailed in Sec.~\ref{sec:5}, our results indicated that these variants underperformed  AlignInstruct when applied to unseen, low-resource languages.

\subsubsection{HintInstruct}
HintInstruct as a generative variant of AlignInstruct was instructions containing word alignment hints. It was inspired by~\citet{DBLP:journals/corr/abs-2302-07856}, where dictionary hints were shown to improve few-shot in-context leaning. Instead of relying on additional dictionaries, we used the same word alignments described in Sec.~\ref{sec:2.2}, which were motivated by the common unavailability of high-quality dictionaries for unseen, low-resource languages. Let $\left\{\left(x_{k_s} x_{k_s+1} \ldots x_{k_s+n_s}, y_{l_s} y_{l_s+1} \ldots y_{l_s+m_s}\right)\right\}^{S}_{s=1}$ be $S$ word pairs extracted from the sentence pair $\left(\left(x_i\right)_1^N,\left(y_j\right)_1^M\right)$. HintInstruct follows the instruction pattern:

\begin{itemize}
    \setlength{\itemsep}{0pt}
    \setlength{\parskip}{0pt}
    \setlength{\topsep}{0pt}
    \item \noindent\textbf{Input}: ``Use the following alignment hints and translate from $Y$ to $X$.

    Alignments between $X$ and $Y$:

    – $\left(x_{k_1} x_{k_1+1} \ldots x_{k_1+n_1}, y_{l_1} y_{l_1+1} \ldots y_{l_1+m_1}\right)$,

    – $\left(x_{k_2} x_{k_2+1} \ldots x_{k_1+n_1}, y_{l_2} y_{l_2+1} \ldots y_{l_2+m_2}\right)$,

    $\ldots$,
    
    – $\left(x_{k_S} x_{k_S+1} \ldots x_{k_S+n_S}, y_{l_S} y_{l_S+1} \ldots y_{l_S+m_S}\right)$, 

    $Y$: $y_1 y_2 \ldots y_M$.

    $X$: ''
    \item \noindent\textbf{Output}: ``$x_1 x_2 \ldots x_N$.''
\end{itemize}
where $S$ denotes the number of the word alignment pairs used to compose the instructions. Different from AlignInstruct, HintInstruct expects the translation targets to be generated.

\subsubsection{ReviseInstruct}
ReviseInstruct was inspired by~\citet{ren-etal-2019-explicit} and~\citet{liu-etal-2020-multilingual-denoising} for the notion of generating parallel words or phrases, thereby encouraging a model to encode cross-lingual alignments. A ReviseInstruct instruction contained a partially corrupted translation target, as well as a directive to identify and revise these erroneous tokens. Tokens are intentionally corrupted at the granularity of individual words, aligning with the word-level granularity in AlignInstruct and HintInstruct. ReviseInstruct follows the instruction pattern:\footnote{We illustrated examples of HintInstruct and ReviseInstruct in App.~\ref{app:hint-revise} for reference.}

\begin{itemize}
    \setlength{\itemsep}{0pt}
    \setlength{\parskip}{0pt}
    \setlength{\topsep}{0pt}
    \item \noindent\textbf{Input}: ``Given the following translation of $X$ from $Y$, output the incorrectly translated word and correct it.

    $Y$: $y_1 y_2 \ldots y_M$.
    
    $X$: $x_1 x_2 \ldots x_k x_{k+1} \ldots x_{k+n} \ldots x_N$.''
    \item \noindent\textbf{Output}: ``The incorrectly translated word is "$x_k x_{k+1} \ldots x_{k+n}$". It should be "$x_j x_{j+1} \ldots x_{j+m}$".''
\end{itemize}


\section{Experimental Settings}

\begin{table*}[t]
    \centering
    \resizebox{\linewidth}{!}{
    \begin{tabular}{cl|rrr|rrr|rrr|rrr}
        \toprule
        BLOOMZ & \multirow{2}{*}{ Objective } & \multicolumn{3}{c}{ OPUS en$\rightarrow$xx } & \multicolumn{3}{|c|}{ OPUS xx$\rightarrow$en } & \multicolumn{3}{c}{ Flores en$\rightarrow$xx } & \multicolumn{3}{|c}{ Flores xx$\rightarrow$en } \\
        model & & BLEU & chrF++ & COMET & BLEU & chrF++ & COMET & BLEU & chrF++ & COMET & BLEU & chrF++ & COMET \\
        \toprule
        \multirow{8}{*}{ BLOOMZ-7b1 } & w/o fine-tuning & 3.61 & 8.82 & 47.81 & 6.70 & 18.49 & 51.68 & 2.00 & 9.35 & 36.54 & 9.95 & 24.47 & 52.05 \\
        \cline{2-14}
        & \multicolumn{10}{l}{ \textit{Individual objectives} } \\
        & MTInstruct & 11.54 & 25.33 & 64.54 & 18.59 & 33.25 & 69.84 & 3.30 & 17.10 & 40.58 & 11.37 & 27.14 & 56.33 \\
        & AlignInstruct & 4.73 & 9.23 & 49.85 & 5.32 & 12.90 & 53.26 & 1.97 & 8.90 & 42.35 & 3.47 & 11.93 & 39.58 \\
        \cline{2-14}
        & \multicolumn{10}{l}{ \textit{Multiple objectives with different curricula} } \\
        & MT+Align & \textbf{12.28} & \textbf{26.17} & \textbf{65.54} & \textbf{18.72} & \textbf{34.02} & \textbf{70.69} & 3.26 & \textbf{17.20} & \textbf{41.07} & \textbf{11.60} & \textbf{27.38} & \textbf{56.98} \\
        & Align$\rightarrow$MT & \textbf{11.73} & \textbf{25.48} & 64.54 & 17.54 & 32.62 & 69.76 & \textbf{3.35} & \textbf{17.21} & \textbf{40.85} & 11.32 & \textbf{27.21} & \textbf{56.50} \\
        & MT+Align$\rightarrow$MT & \textbf{12.10} & \textbf{26.16} & \textbf{65.43} & 18.23 & \textbf{33.54} & \textbf{70.60} & 3.28 & \textbf{17.26} & \textbf{41.13} & \textbf{11.48} & \textbf{27.34} & \textbf{56.78} \\
        \toprule
        \multirow{8}{*}{ BLOOMZ-3b } & w/o fine-tuning & 4.63 & 9.93 & 48.53 & 5.90 & 16.38 & 48.05 & 2.00 & 9.09 & 39.52 & 5.86 & 18.56 & 47.03 \\
        \cline{2-14}
        & \multicolumn{10}{l}{ \textit{Individual objectives} } \\
        & MTInstruct & 10.40 & 23.08 & 62.28 & 16.10 & 31.15 & 68.36 & 2.85 & 16.23 & 39.21 & 8.92 & 24.57 & 53.33 \\
        & AlignInstruct & 1.70 & 4.05 & 43.89 & 0.87 & 3.20 & 41.93 & 0.16 & 3.09 & 31.10 & 0.10 & 1.80 & 29.46 \\
        \cline{2-14}
        & \multicolumn{10}{l}{ \textit{Multiple objectives with different curricula} } \\
        & MT+Align & \textbf{10.61} & \textbf{23.64} & \textbf{62.84} & \textbf{16.73} & \textbf{31.51} & \textbf{68.52} & \textbf{2.95} & \textbf{16.62} & \textbf{39.83} & \textbf{9.50} & \textbf{25.16} & \textbf{54.35} \\
        & Align$\rightarrow$MT & 10.22 & 22.53 & 61.99 & 15.90 & 30.31 & 67.79 & \textbf{3.02} & \textbf{16.43} & \textbf{39.46} & \textbf{9.07} & \textbf{24.70} & \textbf{53.71} \\
        & MT+Align$\rightarrow$MT & \textbf{10.60} & \textbf{23.35} & \textbf{62.69} & \textbf{16.58} & \textbf{31.64} & \textbf{68.98} & \textbf{2.93} & \textbf{16.57} & \textbf{39.78} & \textbf{9.41} & \textbf{25.08} & \textbf{54.13} \\
        \toprule
        \multirow{8}{*}{ BLOOMZ-1b1 } & w/o fine-tuning & 3.76 & 7.57 & 46.98 & 4.78 & 14.11 & 49.34 & 1.24 & 6.93 & 38.13 & 3.49 & 14.56 & 43.26 \\
        \cline{2-14}
        & \multicolumn{10}{l}{ \textit{Individual objectives} } \\
        & MTInstruct & 7.42 & 17.85 & 57.53 & 11.99 & 25.59 & 63.93 & 2.11 & 14.40 & 36.35 & 5.33 & 20.65 & 48.83 \\
        & AlignInstruct & 2.51 & 5.29 & 45.17 & 3.13 & 8.92 & 48.48 & 0.35 & 3.79 & 31.70 & 1.35 & 6.43 & 33.63 \\
        \cline{2-14}
        & \multicolumn{10}{l}{ \textit{Multiple objectives with different curricula} } \\
        & MT+Align & \textbf{7.80} & \textbf{18.48} & \textbf{57.77} & \textbf{12.57} & \textbf{25.92} & \textbf{64.03} & \textbf{2.16} & \textbf{14.54} & \textbf{37.05} & \textbf{5.46} & \textbf{20.90} & \textbf{49.31} \\
        & Align$\rightarrow$MT & \textbf{7.49} & \textbf{18.09} & \textbf{57.67} & 11.80 & 24.70 & 63.29 & 2.08 & 14.28 & \textbf{36.61} & 5.24 & 20.53 & 48.76 \\
        & MT+Align$\rightarrow$MT & \textbf{7.98} & \textbf{18.61} & \textbf{57.94} & \textbf{12.43} & \textbf{25.78} & 63.93 & \textbf{2.16} & \textbf{14.46} & \textbf{37.02} & \textbf{5.37} & \textbf{20.67} & \textbf{49.01} \\
        \bottomrule
    \end{tabular}
    }
    \caption{\textbf{Results of BLOOMZ+24 fine-tuned with MTInstruct and AlignInstruct on different curricula} as described in~\ref{sec:3.2}. Scores that surpass the MTInstruct baseline are marked in \textbf{bold}.}
    \label{tab:res1}
\end{table*}

\subsection{Backbone Models and Unseen Languages}

Our experiments fine-tuned the BLOOMZ models~\cite{muennighoff-etal-2023-crosslingual} for MT in unseen, low-resource languages. BLOOMZ is an instruction fine-tuned multilingual LLM from BLOOM~\cite{DBLP:journals/corr/abs-2211-05100} that supports translation across $46$ languages. 
Two lines of experiments evaluated the effectiveness of the MTInstruct baseline and AlignInstruct:

\noindent\textbf{BLOOMZ+24} Tuning BLOOMZ-7b1, BLOOMZ-3b, and BLOOMZ-1b1\footnote{\url{https://huggingface.co/bigscience/bloomz}} for $24$ unseen, low-resource languages. These experiments aimed to: (1) assess the effectiveness of AlignInstruct in multilingual, low-resource scenarios; (2) offer comparison across various model sizes. We used the OPUS-100~\cite{zhang-etal-2020-improving}\footnote{\url{https://opus.nlpl.eu/opus-100.php}} datasets as training data. OPUS-100 is an English-centric parallel corpora, with around $4.5$M parallel sentences in total for 24 selected languages, averaging $187$k sentence pairs for each language and English. Refer to App.~\ref{app:data} for training data statistics. We used OPUS-100 and Flores-200~\cite{DBLP:journals/corr/abs-2207-04672}\footnote{\url{https://github.com/facebookresearch/flores/blob/main/flores200/README.md}} for evaluating translation between English and $24$ unseen languages (48 directions in total) on in-domain and out-of-domain test sets, respectively. The identical prompt as introduced in Sec.~\ref{sec:2.1} was employed for inference. Inferences using alternative MT prompts are discussed in App.~\ref{app:prompts}.


\noindent\textbf{BLOOMZ+3} Tuning BLOOMZ-7b1 with three unseen languages, German, Dutch, and Russian, or a combination of these three unseen languages and another three seen (Arabic, French, and Chinese). We denote the respective setting as \textbf{de-nl-ru} and \textbf{ar-de-fr-nl-ru-zh}. These experiments assessed the efficacy of AlignInstruct in zero-shot translation scenarios, where translation directions were not presented during fine-tuning, as well as the translation performance when incorporating supported languages as either source or target languages. 
To simulate the low-resource fine-tuning scenario, we randomly sampled $200$k parallel sentences for each language. For evaluation, we used the OPUS-100 supervised and zero-shot test sets, comprising 12 supervised directions involving English and 30 zero-shot directions without English among six languages.


Notably, BLOOMZ's pre-training data includes the English portion of the Flores-200 dataset, potentially leading to data leakage during evaluation~\cite{muennighoff-etal-2023-crosslingual,DBLP:journals/corr/abs-2304-04675}. To mitigate this, our evaluation also compared translation quality before and after fine-tuning, thereby distinguishing the genuine improvements in translation capability attributable to the fine-tuning process (refer to the results in Sec.~\ref{sec:5}).

\subsection{Training Details and Curricula}
\label{sec:3.2}
The PEFT method, LoRA~\cite{DBLP:conf/iclr/HuSWALWWC22}, was chosen to satisfy the parameter efficiency requirement for low-resource languages, as full-parameter fine-tuning would likely under-specify the models.See App.~\ref{app:training} for implementation details. How AlignInstruct and MTInstruct are integrated into training remained undetermined. To that end, we investigated three training curricula:

\begin{table*}[t]
    \centering
    \resizebox{0.94\linewidth}{!}{
    \begin{tabular}{l|rr|rr|rr|rr|rr|rr}
    \toprule
    Objective & en-\textbf{af} & \textbf{af}-en & en-\textbf{am} & \textbf{am}-en & en-\textbf{be} & \textbf{be}-en & en-\textbf{cy} & \textbf{cy}-en & en-\textbf{ga} & \textbf{ga}-en & en-\textbf{gd} & \textbf{gd}-en \\
    \hline
    MTInstruct & 25.0 & 38.5 & 3.0 & 3.4 & 8.9 & 14.0 & 20.2 & 33.2 & 15.6 & 29.2 & 13.1 & 66.0 \\
    MT+Align   & 25.0 & \textit{36.9} & \textbf{3.4} & \textbf{4.9} & \textit{8.3} & 13.9 & \textbf{20.6} & \textbf{33.8} & \textbf{17.6} & \textbf{32.6} & \textbf{15.6} & \textit{48.1} \\
    \toprule
    Objective & en-\textbf{gl} & \textbf{gl}-en & en-\textbf{ha} & \textbf{ha}-en & en-\textbf{ka} & \textbf{ka}-en & en-\textbf{kk} & \textbf{kk}-en & en-\textbf{km} & \textbf{km}-en & en-\textbf{ky} & \textbf{ky}-en \\
    \hline
    MTInstruct & 16.9 & 24.7 & 12.3 & 10.0 & 4.6 & 10.0 & 12.6 & 14.6 & 19.7 & 13.9 & 16.0 & 21.1 \\
    MT+Align   & 17.1 & 24.4 & \textbf{14.6} & \textbf{11.4} & 4.9 & \textbf{10.5} & 12.3 & \textbf{15.6} & \textbf{20.4} & \textbf{14.4} & 15.8 & \textbf{23.3} \\
    \toprule
    Objective & en-\textbf{li} & \textbf{li}-en & en-\textbf{my} & \textbf{my}-en & en-\textbf{nb} & \textbf{nb}-en & en-\textbf{nn} & \textbf{nn}-en & en-\textbf{oc} & \textbf{oc}-en & en-\textbf{si} & \textbf{si}-en \\
    \hline
    MTInstruct & 13.5 & 21.3 & 6.2 & 5.2 & 12.7 & 22.2 & 18.3 & 27.1 & 10.0 & 13.4 & 5.2 & 11.5 \\
    MT+Align   & 13.2 & \textbf{22.3} & \textbf{7.6} & \textbf{6.3} & \textbf{13.5} & \textbf{24.2} & \textbf{19.0} & \textbf{28.5} & \textit{9.1} & 13.5 & 5.1 & \textbf{13.9} \\
    \toprule
    Objective & en-\textbf{tg} & \textbf{tg}-en & en-\textbf{tk} & \textbf{tk}-en & en-\textbf{tt} & \textbf{tt}-en & en-\textbf{ug} & \textbf{ug}-en & en-\textbf{uz} & \textbf{uz}-en & en-\textbf{yi} & \textbf{yi}-en \\
    \hline
    MTInstruct & 5.5 & 8.0 & 24.4 & 30.4 & 1.9 & 3.6 & 1.2 & 4.2 & 3.1 & 5.7 & 7.1 & 14.9 \\
    MT+Align   & \textbf{6.6} & \textbf{8.8} & \textbf{27.2} & \textbf{31.2} & 2.1 & \textbf{5.0} & 1.1 & \textbf{5.5} & \textbf{3.5} & \textbf{7.4} & \textbf{11.1} & \textit{12.8} \\
    \bottomrule
    \end{tabular}
    }
    \caption{Language-wise BLEU results on BLOOMZ-7b1 for BLOOMZ+24 fine-tuned using MTInstruct or MT+Align. Scores significantly~\cite{koehn-2004-statistical} outperforming the MTInstruct baseline are emphasized in \textbf{bold} while those decreased significantly~\cite{koehn-2004-statistical} are marked in \textit{italics}.}
    \label{bloom+24}
\end{table*}

\begin{table*}[t]
    \centering
    \resizebox{0.975\linewidth}{!}{
    \begin{tabular}{cl|rrr|rrr|rrr|rrr}
        \toprule
        BLOOMZ & \multirow{2}{*}{ Objective } & \multicolumn{3}{c|}{ OPUS en$\rightarrow$xx } & \multicolumn{3}{c|}{ OPUS xx$\rightarrow$en } & \multicolumn{3}{c|}{ Flores en$\rightarrow$xx } & \multicolumn{3}{c}{ Flores xx$\rightarrow$en } \\
        model & & BLEU & chrF++ & COMET & BLEU & chrF++ & COMET & BLEU & chrF++ & COMET & BLEU & chrF++ & COMET \\
        \toprule
        \multirow{5}{*}{ BLOOMZ-7b1} & MTInstruct & 11.54 & 25.33 & 64.54 & 18.59 & 33.25 & 69.84 & 3.30 & 17.10 & 40.58 & 11.37 & 27.14 & 56.33 \\
        \cline{2-14}
        & MT+Align & \textbf{12.28} & \textbf{26.17} & \textbf{65.54} & \textbf{18.72} & \textbf{34.02} & \textbf{70.69} & 3.26 & \textbf{17.20} & \textbf{41.07} & \textbf{11.60} & \textbf{27.38} & \textbf{56.98} \\
        & MT+Hint & \textbf{12.12} & \textbf{25.92} & \textbf{64.60} & 18.25 & 33.18 & \textbf{70.31} & \textbf{3.34} & \textbf{17.13} & \textbf{41.10} & \textbf{11.45} & \textbf{27.37} & \textbf{56.86} \\
        & MT+Revise & \textbf{11.96} & \textbf{25.73} & \textbf{64.73} & \textbf{18.69} & \textbf{33.74} & \textbf{70.32} & \textbf{3.34} & 17.10 & \textbf{41.07} & \textbf{11.44} & \textbf{27.37} & \textbf{56.73} \\
        \toprule
        \multirow{5}{*}{ BLOOMZ-3b} & MTInstruct & 10.40 & 23.08 & 62.28 & 16.10 & 31.15 & 68.36 & 2.85 & 16.23 & 39.21 & 8.92 & 24.57 & 53.33 \\
        \cline{2-14}
        & MT+Align & \textbf{10.61} & \textbf{23.64} & \textbf{62.84} & \textbf{16.73} & \textbf{31.51} & \textbf{68.52} & \textbf{2.95} & \textbf{16.62} & \textbf{39.83} & \textbf{9.50} & \textbf{25.16} & \textbf{54.35} \\
        & MT+Hint & \textbf{10.49} & \textbf{23.34} & \textbf{62.65} & \textbf{16.29} & \textbf{31.43} & \textbf{68.83} & \textbf{3.11} & \textbf{16.95} & \textbf{39.91} & \textbf{9.52} & \textbf{25.25} & \textbf{54.28} \\
        & MT+Revise & \textbf{10.52} & 23.03 & 62.04 & \textbf{16.22} & 30.98 & 68.28 & \textbf{2.99} & \textbf{16.83} & \textbf{39.52} & \textbf{9.47} & \textbf{25.21} & \textbf{53.91} \\
        \toprule
        \multirow{5}{*}{ BLOOMZ-1b1 } & MTInstruct & 7.42 & 17.85 & 57.53 & 11.99 & 25.59 & 63.93 & 2.11 & 14.40 & 36.35 & 5.33 & 20.65 & 48.83 \\
        \cline{2-14}
        & MT+Align & \textbf{7.80} & \textbf{18.48} & \textbf{57.77} & \textbf{12.57} & \textbf{25.92} & \textbf{64.03} & \textbf{2.16} & \textbf{14.54} & \textbf{37.05} & \textbf{5.46} & \textbf{20.90} & \textbf{49.31} \\
        & MT+Hint & \textbf{7.71} & \textbf{18.15} & \textbf{57.76} & 11.52 & 24.88 & 63.63 & \textbf{2.21} & \textbf{14.61} & \textbf{37.24} & \textbf{5.47} & \textbf{20.78} & \textbf{48.97} \\
        & MT+Revise & 7.31 & \textbf{17.99} & 57.45 & \textbf{12.00} & 25.33 & 63.81 & 2.07 & 14.32 & \textbf{36.68} & \textbf{5.41} & \textbf{20.91} & \textbf{49.09} \\
        \bottomrule
    \end{tabular}
    }
    \caption{\textbf{Results of BLOOMZ+24 fine-tuned combining MTInstruct with AlignInstruct (or its generative variants).} Scores that surpass the MTInstruct baseline are marked in \textbf{bold}.}
    \label{tab:res2}
\end{table*}

\begin{table*}[t]
    \centering
    \resizebox{0.975\linewidth}{!}{
    \begin{tabular}{l|rrr|rrr|rrr|rrr}
        \toprule
        \multirow{2}{*}{ Objective } & \multicolumn{3}{c|}{ OPUS en$\rightarrow$xx } & \multicolumn{3}{c|}{ OPUS xx$\rightarrow$en } & \multicolumn{3}{c|}{ Flores en$\rightarrow$xx } & \multicolumn{3}{c}{ Flores xx$\rightarrow$en } \\
        & BLEU & chrF++ & COMET & BLEU & chrF++ & COMET & BLEU & chrF++ & COMET & BLEU & chrF++ & COMET \\
        \toprule
        MTInstruct & 11.54 & 25.33 & 64.54 & 18.59 & 33.25 & 69.84 & 3.30 & 17.10 & 40.58 & 11.37 & 27.14 & 56.33 \\
        MT+Align & \textbf{12.28} & \textbf{26.17} & \textbf{65.54} & \textbf{18.72} & \textbf{34.02} & \textbf{70.69} & 3.26 & \textbf{17.20} & \textbf{41.07} & \textbf{11.60} & \textbf{27.38} & \textbf{56.98} \\
        \hline
        MT+Align+Revise & \textbf{12.08} & \textbf{25.73} & \textbf{64.55} & \textbf{19.23} & \textbf{34.32} & \textbf{70.60} & \textbf{3.33} & \textbf{17.25} & \textbf{41.17} & \textbf{11.60} & \textbf{27.61} & \textbf{57.22} \\
        MT+Align+Hint & \textbf{12.02} & \textbf{25.51} & \textbf{64.58} & \textbf{19.40} & \textbf{34.44} & \textbf{70.65} & 3.25 & 16.87 & \textbf{41.13} & \textbf{11.58} & \textbf{27.48} & \textbf{56.93} \\
        MT+Hint+Revise & \textbf{12.10} & \textbf{25.69} & \textbf{64.68} & \textbf{19.58} & \textbf{34.49} & \textbf{70.55} & \textbf{3.34} & \textbf{17.24} & \textbf{41.13} & \textbf{11.70} & \textbf{27.62} & \textbf{57.19} \\
        MT+Align+Hint+Revise & \textbf{12.00} & \textbf{25.39} & \textbf{64.55} & \textbf{19.68} & \textbf{34.48} & \textbf{70.64} & \textbf{3.40} & \textbf{17.17} & \textbf{41.21} & \textbf{11.67} & \textbf{27.54} & \textbf{57.16} \\
        \toprule
    \end{tabular}
    }
    \caption{\textbf{Results of BLOOMZ+24 combining MTInstruct with multiple objectives among AlignInstruct, HintInstruct, and ReviseInstruct on BLOOMZ-7b1.} Scores that surpass MTInstruct are marked in \textbf{bold}.}
    \label{tab:res3}
\end{table*}

\noindent\textbf{Multi-task Fine-tuning} 
combined multiple tasks in a single training session~\cite{multi-task-learning}. This was realized by joining MTInstruct and AlignInstruct training data, denoted as \textbf{MT+Align}.\footnote{Note that AlignInstruct and MTInstruct were derived from the same parallel corpora.}



\noindent\textbf{Pre-fine-tuning \& Fine-tuning} 
arranges fine-tuning in a two-stage curriculum~\cite{DBLP:conf/icml/BengioLCW09}, first with AlignInstruct, then with MTInstruct.\footnote{An effective curriculum often starts with a simple and general task, followed by a task-specific task.} This configuration, denoted as \textbf{Align$\rightarrow$MT}, validates whether AlignInstruct should precede MTInstruct.


\noindent\textbf{Mixed Fine-tuning}
~\cite{chu-etal-2017-empirical} arranged the above curricula to start with MT+Align, followed by MTInstruct, denoted as \textbf{MT+Align$\rightarrow$MT}.


\begin{table*}[t]
    \centering
    \resizebox{0.84\linewidth}{!}{
    \begin{tabular}{ll|lrrr|lrrr}
        \toprule
        Fine-tuned & \multirow{2}{*}{ Objective } & \multicolumn{4}{c|}{ Zero-shot Directions } & \multicolumn{4}{c}{ Supervised Directions } \\
        Languages & & Directions & BLEU & chrF++ & COMET & Directions & BLEU & chrF++ & COMET \\
        \toprule
        \multirow{6}{*}{-} & \multirow{6}{*}{w/o fine-tuning} & \multirow{2}{*}{overall} & \multirow{2}{*}{6.89} & \multirow{2}{*}{19.14} & \multirow{2}{*}{57.95} & en$\rightarrow$xx & 13.38 & 26.65 & 64.28 \\
        & & & & & & xx$\rightarrow$en & 21.70 & 42.05 & 72.72 \\
        & & seen$\rightarrow$seen & 16.95 & 30.78 & 74.58 & en$\rightarrow$seen & 20.13 & 32.87 & 76.99 \\
        & & seen$\rightarrow$unseen & 2.30 & 13.31 & 49.98 & en$\rightarrow$unseen & 6.63 & 20.43 & 51.56 \\
        & & unseen$\rightarrow$seen & 7.78 & 20.07 & 62.74 & seen$\rightarrow$en & 26.30 & 48.70 & 78.22 \\
        & & unseen$\rightarrow$unseen & 2.37 & 14.83 & 46.06 & unseen$\rightarrow$en & 17.10 & 35.40 & 67.23 \\
        \toprule
        \multirow{12}{*}{ de-nl-ru } & \multirow{6}{*}{ MTInstruct } & \multirow{2}{*}{overall} & \multirow{2}{*}{8.38} & \multirow{2}{*}{22.75} & \multirow{2}{*}{59.93} & en$\rightarrow$xx & 17.05 & 32.02 & 69.26 \\
        & & & & & & xx$\rightarrow$en & 25.13 & 45.02 & 76.29 \\
        & & seen$\rightarrow$seen & 14.52 & 27.25 & 70.48 & en$\rightarrow$seen & 17.60 & 29.87 & 73.81 \\
        & & seen$\rightarrow$unseen & 6.14 & 22.82 & 54.75 & en$\rightarrow$unseen & 16.50 & 34.17 & 64.70 \\
        & & unseen$\rightarrow$seen & 7.56 & 19.22 & 61.99 & seen$\rightarrow$en & 25.73 & 47.07 & 77.52 \\
        & & unseen$\rightarrow$unseen & 6.85 & 23.45 & 54.07 & unseen$\rightarrow$en & 24.53 & 42.97 & 75.06 \\
        \cline{2-10}
        & \multirow{6}{*}{MT+Align} & \multirow{2}{*}{overall} & \multirow{2}{*}{\textbf{8.86}} & \multirow{2}{*}{\textbf{23.30}} & \multirow{2}{*}{\textbf{60.70}} & en$\rightarrow$xx & 16.63 & 31.73 & 68.79 \\
        & & & & & & xx$\rightarrow$en & \textbf{25.62} & \textbf{45.37} & \textbf{76.45} \\
        & & seen$\rightarrow$seen & \textbf{14.77} & \textbf{27.80} & \textbf{71.07} & en$\rightarrow$seen & 15.80 & 28.47 & 72.35 \\
        & & seen$\rightarrow$unseen & \textbf{6.31} & \textbf{23.08} & \textbf{54.81} & en$\rightarrow$unseen & \textbf{17.47} & \textbf{35.00} & \textbf{65.24} \\
        & & unseen$\rightarrow$seen & \textbf{8.61} & \textbf{20.24} & \textbf{63.81} & seen$\rightarrow$en & \textbf{25.90} & \textbf{47.13} & 77.47 \\
        & & unseen$\rightarrow$unseen & \textbf{7.15} & \textbf{23.70} & \textbf{54.51} & unseen$\rightarrow$en & \textbf{25.33} & \textbf{43.60} & \textbf{75.43} \\
        \bottomrule
        \multirow{12}{*}{ ar-de-fr-nl-ru-zh } & \multirow{6}{*}{ MTInstruct } & \multirow{2}{*}{overall} & \multirow{2}{*}{11.79} & \multirow{2}{*}{26.36} & \multirow{2}{*}{63.22} & en$\rightarrow$xx & 21.18 & 35.52 & 70.86 \\
        & & & & & & xx$\rightarrow$en & 28.35 & 48.00 & 77.30 \\
        & & seen$\rightarrow$seen & 22.68 & 35.32 & 76.39 & en$\rightarrow$seen & 26.20 & 37.77 & 78.22 \\
        & & seen$\rightarrow$unseen & 7.10 & 24.50 & 55.18 & en$\rightarrow$unseen & 16.17 & 33.27 & 63.50 \\
        & & unseen$\rightarrow$seen & 12.56 & 24.74 & 68.83 & seen$\rightarrow$en & 31.97 & 52.93 & 79.72 \\
        & & unseen$\rightarrow$unseen & 6.78 & 22.62 & 53.69 & unseen$\rightarrow$en & 24.73 & 43.07 & 74.88 \\
        \cline{2-10}
        & \multirow{6}{*}{ MT+Align } & \multirow{2}{*}{overall} & \multirow{2}{*}{\textbf{12.13}} & \multirow{2}{*}{\textbf{26.65}} & \multirow{2}{*}{\textbf{63.23}} & en$\rightarrow$xx & \textbf{21.33} & \textbf{35.65} & \textbf{70.99} \\
        & & & & & & xx$\rightarrow$en & \textbf{28.60} & \textbf{48.27} & \textbf{77.49} \\
        & & seen$\rightarrow$seen & \textbf{23.67} & \textbf{36.53} & \textbf{76.89} & en$\rightarrow$seen & \textbf{26.30} & 37.63 & \textbf{78.25} \\
        & & seen$\rightarrow$unseen & \textbf{7.27} & 24.32 & 54.96 & en$\rightarrow$unseen & \textbf{16.37} & \textbf{33.67} & \textbf{63.73} \\
        & & unseen$\rightarrow$seen & \textbf{12.92} & \textbf{25.29} & \textbf{69.10} & seen$\rightarrow$en & \textbf{32.03} & \textbf{53.07} & \textbf{79.93} \\
        & & unseen$\rightarrow$unseen & 6.68 & 22.30 & 53.19 & unseen$\rightarrow$en & \textbf{25.17} & \textbf{43.47} & \textbf{75.05} \\
        \bottomrule
    \end{tabular}
    }
    \caption{\textbf{Results of BLOOMZ+3 without fine-tuning or fine-tuned with MTInstruct, or MT+Align.} Scores that surpass the MTInstruct baseline are marked in \textbf{bold}. ``Seen'' and ``unseen'' refer to whether the language was included in the pre-training of the BLOOMZ model. xx includes seen and unseen languages.}
    \label{tab:res5}
\end{table*}

\section{Evaluation and Analysis}
\label{sec:5}
This section reports BLEU~\cite{papineni-etal-2002-bleu,post-2018-call}, chrF++~\cite{popovic-2015-chrf}, and COMET~\cite{rei-etal-2020-comet}\footnote{COMET scores do not currently support Limburgish (li), Occitan (oc), Tajik (tg), Turkmen (tk), and Tatar (tt) among the 24 languages in the BLOOMZ+24 setting. Thus, we report the average COMET scores for the remaining 19 languages.} scores for respective experimental configurations. We further characterized of the degree to which intermediate embeddings were language-agnostic after fine-tuning.

\subsection{BLOOMZ+24 Results}

Tab.~\ref{tab:res1} shows the scores for the unmodified BLOOMZ models, as well as BLOOMZ+24 under MTInstruct, AlignInstruct, and the three distinct curricula. Non-trivial improvements in all metrics were evident for BLOOMZ+24 under MTInstruct. This suggests that MTInstruct can induce translation capabilities in unseen languages. Applying AlignInstruct and MTInstruct via the curricula further showed better scores than the baselines, suggesting the role of AlignInstruct as complementing MTInstruct. Align$\rightarrow$MT was an exception, performing similarly to MTInstruct. This may indicate the effect of AlignInstruct depends on its cadence relative to MTInstruct in a curriculum. 

Superior OPUS and Flores scores under the xx$\rightarrow$en direction were evident, compared to the reverse direction, en$\rightarrow$xx. This suggests that our treatments induced understanding capabilities more than generative ones. This may be attributed to the fact that BLOOMZ had significant exposure to English, and that we used English-centric corpora. Finally, we noted the inferior performance of Flores than OPUS. This speaks to the challenge of instilling out-of-domain translation abilities in unseen languages. Our future work will focus on enhancing the domain generalization capabilities of LLM fine-tuning in MT tasks.

Moreover, we reported the language-wise scores in Tab.~\ref{bloom+24}. Specifically, in the ``en-xx'' direction, 11 languages showed statistically significant~\cite{koehn-2004-statistical} improvements, and only 2 decreased significantly. In the ``xx-en'' direction, the improvements were more pronounced, with 18 languages improving significantly (most by over 1 BLEU point) and 3 decreasing significantly. The average improvement for ``en-xx'' was 0.74, which was substantial, especially given the limited volume of parallel data available for each language. The smaller average increase in ``xx-en'' can be attributed to a large decrease in one language (gd), likely due to limited training data (which can be potentially addressed with oversampling). The significantly enhanced performance in most individual languages underscores the effectiveness of our proposed methods.

\subsection{Assessing AlignInstruct Variants}
From Tab.~\ref{tab:res2}, we observed the objectives with AlignInstruct consistently outperformed those with HintInstruct or ReviseInstruct across metrics and model sizes. Namely, easy, discriminative instructions, rather than hard, generative ones, may be preferred for experiments under similar data constraints. The low-resource constraint likely made MTInstruct more sensitive to the difficulty of its accompanying tasks. 

Further, combining more than two instruction tuning tasks simultaneously did not guarantee consistent improvements, see Tab.~\ref{tab:res3}. Notably, MT+Align either outperformed or matched the performance of other objective configurations. While merging multiple instruction tuning tasks occasionally resulted in superior BLEU and chrF++ scores for OPUS xx$\rightarrow$en, it fell short in COMET scores compared to MT+Align. This indicated that while such configurations might enhance word-level translation quality, as reflected by BLEU and chrF++ scores, due to increased exposure to cross-lingual word alignments, MT+Align better captured the context of the source sentence as reflected by COMET scores. Overall, these instruction tuning tasks did not demonstrate significant synergistic effects for fine-tuning for unseen languages.


\subsection{BLOOMZ+3 Zero-shot Evaluation}
Tab.~\ref{tab:res5} reports the results of the two settings, de-nl-ru and ar-de-fr-nl-ru-zh. Results of MT+Align+Hint+Revise and pivot-based translation are reported in App.~\ref{app:zero} and~\ref{app:pivot}. In the de-nl-ru setting, where BLOOMZ was fine-tuned with the three unseen languages, we noticed MT+Align consistently outperformed the MTInstruct baseline across all evaluated zero-shot directions. Notably, MT+Align enhanced the translation quality for unseen$\rightarrow$seen and seen$\rightarrow$unseen directions compared to w/o fine-tuning and MTInstruct, given that the model was solely fine-tuned on de, nl, and ru data. This suggested AlignInstruct not only benefits the languages supplied in the data but also has a positive impact on other languages through cross-lingual alignment supervision. In terms of supervised directions involving English, we noticed performance improvements associated with unseen languages, and regression in seen ones. The regression may be attributed to forgetting for the absence of seen languages in fine-tuning data. Indeed, continuous exposure to English maintained the translation quality for seen$\rightarrow$en. As LoRA is modular, the regression can be mitigated by detaching the LoRA parameters for seen languages.

The ar-de-fr-nl-ru-zh setting yielded a consistently higher translation quality across all directions when compared with the de-nl-ru setting. This improvement was expected, as all the six languages were included. Translation quality improved for when generating seen languages under the zero-shot scenario. However, the same observation cannot be made for unseen languages. This phenomenon underscored the effectiveness of AlignInstruct in enhancing translation quality for BLOOMZ's supported languages, but suggested limitations for unseen languages when mixed with supported languages in zero-shot scenarios. In the supervised directions, we found all translation directions surpassed the performance of the MTInstruct baseline. This highlighted the overall effectiveness of AlignInstruct in enhancing translation quality across a range of supervised directions. 

\begin{figure}
    \centering
    \includegraphics[width=\linewidth]{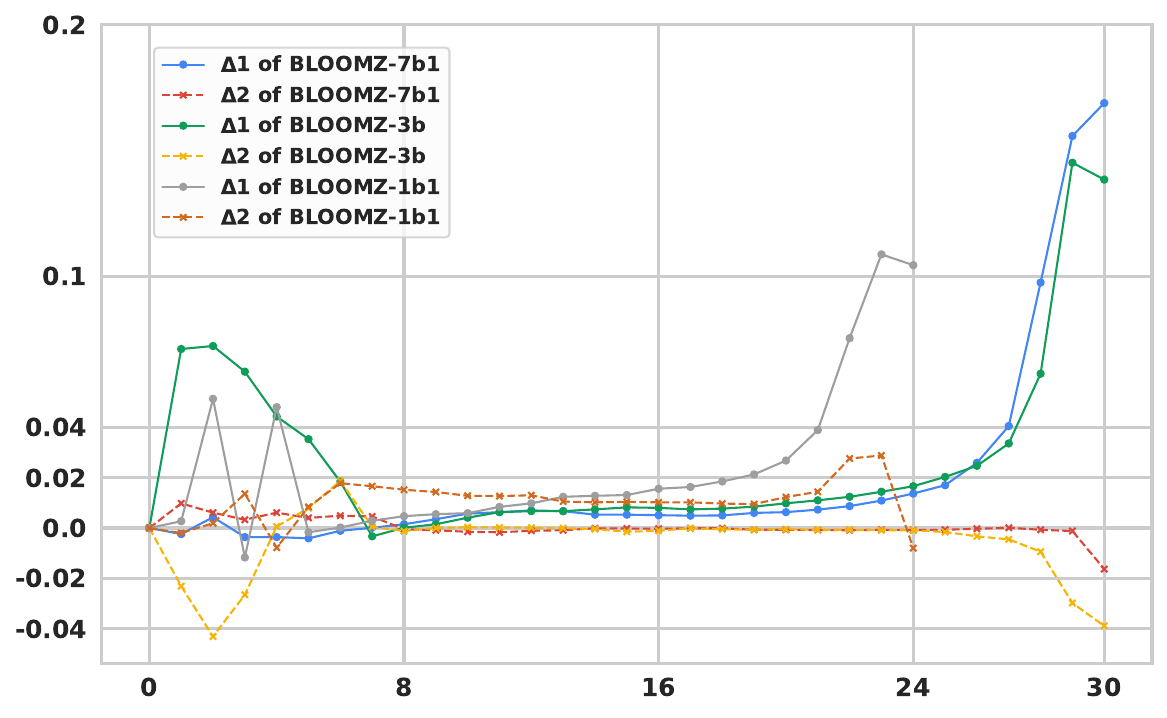}
    \caption{\textbf{Differences in cosine similarity of layer-wise embeddings for BLOOMZ+24.} $\Delta$1 represents the changes from the unmodified BLOOMZ to the one on MTInstruct, and $\Delta$2 from MTInstruct to MT+Align.}
    \label{fig:cosine}
\end{figure}

\subsection{How did MTInstruct and AlignInstruct Impact BLOOMZ's Representations?}
\label{sec:4.5}

This section analyzed the layer-wise cosine similarities between the embeddings of parallel sentences to understand the changes in internal representations after fine-tuning. The parallel sentences were prepared from the English-centric validation datasets. We then mean-pool the outputs at each layer as sentence embeddings and compute the cosine similarities, as illustrated in Fig.~\ref{fig:cosine}. Results for BLOOMZ+3 are discussed in App.~\ref{app:embedding}.

We observed that, after MTInstruct fine-tuning, the cosine similarities rose in nearly all layers ($\Delta$1, Fig.~\ref{fig:cosine}). This may be interpreted as enhanced cross-lingual alignment, and as indicating the acquisition of translation capabilities. Upon further combination with AlignInstruct ($\Delta$2, Fig.~\ref{fig:cosine}), the degree of cross-lingual alignment rose in the early layers (layers 4 - 7) then diminished in the final layers (layers 29 \& 30). This pattern aligned with the characteristics of encoder-decoder multilingual NMT models, where language-agnostic encoder representations with language-specific decoder representations improve multilingual NMT performance~\cite{liu-etal-2021-improving-zero,wu-etal-2021-language,mao-etal-2023-exploring}. This highlights the beneficial impact of AlignInstruct.


\section{Related Work}

\noindent \textbf{Prompting LLMs for MT}
LLMs have shown good performance for multilingual MT through few-shot in-context learning (ICL)~\cite{DBLP:journals/corr/abs-2301-08745}.~\citet{agrawal-etal-2023-context} and~\citet{DBLP:conf/icml/0006HB23} explored strategies to compose better examples for ICL for XGLM-7.5B~\cite{lin-etal-2022-shot} and GLM-130B~\cite{DBLP:conf/iclr/ZengLDWL0YXZXTM23}.~\citet{DBLP:journals/corr/abs-2302-07856},~\citet{DBLP:journals/corr/abs-2303-13780}, and~\citet{moslem-etal-2023-adaptive} claimed that dictionary-based hints and domain-specific style information can improve prompting OPT~\cite{DBLP:journals/corr/abs-2205-01068}, GPT-3.5~\cite{DBLP:conf/nips/BrownMRSKDNSSAA20}, and BLOOM~\cite{DBLP:journals/corr/abs-2211-05100} for MT.~\citet{DBLP:journals/corr/abs-2305-04118} used LLMs to mine useful knowledge for prompting GPT-3.5 for MT.



\noindent \textbf{Fine-tuning LLMs for MT}
ICL-based methods do not support languages unseen during pre-training. Current  approaches address this issue via fine-tuning.~\citet{DBLP:journals/corr/abs-2306-10968} explored adding new languages to LLaMA~\cite{DBLP:journals/corr/abs-2302-13971} with interactive translation task for unseen high-resource languages. However, similar task datasets are usually not available for most unseen, low-resource languages.~\citet{DBLP:journals/corr/abs-2305-15083} and~\citet{DBLP:journals/corr/abs-2309-11674} showed multilingual fine-tuning with translation instructions can improve the translation ability in supported languages. Our study extended their finding to apply in the context of unseen, low-resource languages. In parallel research,~\citet{DBLP:journals/corr/abs-2305-18098} undertook MT instruction fine-tuning in a massively multilingual context for unseen languages. However, their emphasis was on fine-tuning curriculum based on resource availability of languages, whereas we exclusively centered on low-resource languages and instruction tuning tasks.



\section{Conclusion}
In this study, we introduced AlignInstruct for enhancing the fine-tuning of LLMs for MT in unseen, low-resource languages while limiting the use of additional training corpora. Our multilingual and zero-shot findings demonstrated the strength of AlignInstruct over the MTInstruct baseline and other instruction variants. Our future work pertains to exploring using large monolingual corpora of unseen languages for MT and refining the model capability to generalize across diverse MT prompts.

\section*{Limitations}

\noindent \textbf{Multilingual LLMs}
In this study, our investigations were confined to the fine-tuning of BLOOMZ models with sizes of 1.1B, 3B, and 7.1B. We did not experiment with the 175B BLOOMZ model due to computational resource constraints. However, examining this model could provide valuable insights into the efficacy of our proposed techniques. Additionally, it would be instructive to experiment with other recent open-source multilingual LLMs, such as mGPT~\cite{DBLP:journals/corr/abs-2204-07580} and LLaMa2~\cite{DBLP:journals/corr/abs-2307-09288}.

\noindent \textbf{PEFT Methods and Adapters}
As discussed in the BLOOM+1 paper~\cite{yong-etal-2023-bloom}, alternative PEFT techniques, such as (IA)$^3$~\cite{DBLP:conf/nips/LiuTMMHBR22}, have the potential to enhance the adaptation performance of LLM pre-training for previously unseen languages. These approaches are worth exploring for MT fine-tuning in such languages, in addition to the LoRA methods employed in this study. Furthermore, our exploration was limited to fine-tuning multiple languages using shared additional parameters. Investigating efficient adaptation through the use of the mixture of experts (MoE) approach for MT tasks~\cite{DBLP:journals/jmlr/FanBSMEGBCWCGBL21,DBLP:journals/corr/abs-2207-04672,mohammadshahi-etal-2022-small,koishekenov-etal-2023-memory,DBLP:journals/corr/abs-2305-13993} presents another intriguing avenue for LLM fine-tuning.

\noindent \textbf{Instruction Fine-tuning Data}
Another limitation of our study is that we exclusively explored MT instruction fine-tuning using fixed templates to create MT and alignment instructions. Investigating varied templates (either manually~\cite{DBLP:journals/corr/abs-2305-18098} or automatically constructed~\cite{DBLP:conf/iclr/ZhouMHPPCB23}) might enhance the fine-tuned MT model's ability to generalize across different MT task descriptions. Additionally, leveraging large monolingual corpora in unseen languages could potentially enhance the effectiveness of monolingual instructions for MT downstream tasks, offering further insights beyond the resource-constrained scenarios examined in this work. Furthermore, the creation and utilization of instruction tuning datasets, akin to xP3~\cite{muennighoff-etal-2023-crosslingual}, for unseen, low-resource languages could potentially amplify LLMs' proficiency in following instructions in such languages.~\citet{DBLP:journals/corr/abs-2308-04948} has investigated multilingual instruction tuning datasets. However, the scalability of such high-quality datasets to thousands of low-resource languages still remains to be addressed.

\noindent \textbf{Comparison with the State-of-the-art Multilingual NMT Models}
In this study, we refrained from contrasting translations in low-resource languages with best-performing multilingual NMT models like NLLB-200~\cite{DBLP:journals/corr/abs-2207-04672}, as our primary objective centered on enhancing the MTInstruct baseline through improved cross-lingual alignment within LLMs, rather than delving into the best combination of techniques for MT fine-tuning in LLMs. In future exploration, our methods can potentially be integrated with the MT fine-tuning paradigm proposed by the concurrent work of~\citet{DBLP:journals/corr/abs-2309-11674}, paving the way for elevating the state-of-the-art translation quality using LLMs.


\bibliography{anthology,custom}

\begin{table*}[t!]
    \centering
    \resizebox{0.9\linewidth}{!}{
    \begin{tabular}{lllllcr}
        \toprule
        Language & ISO 639-1 & Language Family & Subgrouping & Script & Seen Script & \#sent. \\
        \toprule
        Afrikaans & af & Indo-European & Germanic & Latin & \Checkmark & 275,512 \\
        Amharic & am & Afro-Asiatic & Semitic & Ge'ez & \XSolidBrush & 89,027 \\
        Belarusian & be & Indo-European & Balto-Slavic & Cyrillic & \XSolidBrush & 67,312 \\
        Welsh & cy & Indo-European & Celtic & Latin & \Checkmark & 289,521 \\
        Irish & ga & Indo-European & Celtic & Latin & \Checkmark & 289,524 \\
        Scottish Gaelic & gd & Indo-European & Celtic & Latin & \Checkmark & 16,316 \\
        Galician & gl & Indo-European & Italic & Latin & \Checkmark & 515,344 \\
        Hausa & ha & Afro-Asiatic & Chadic & Latin & \Checkmark & 97,983 \\
        Georgian & ka & Kartvelian & Georgian-Zan & Georgian & \XSolidBrush & 377,306 \\
        Kazakh & kk & Turkic & Common Turkic & Cyrillic & \XSolidBrush & 79,927 \\
        Khmer & km & Austroasiatic & Khmeric & Khmer & \XSolidBrush & 111,483 \\
        Kyrgyz & ky & Turkic & Common Turkic & Cyrillic & \XSolidBrush & 27,215 \\
        Limburgish & li & Indo-European & Germanic & Latin & \Checkmark & 25,535 \\
        Burmese & my & Sino-Tibetan & Burmo-Qiangic & Myanmar & \XSolidBrush & 24,594 \\
        Norwegian Bokmål & nb & Indo-European & Germanic & Latin & \Checkmark & 142,906 \\
        Norwegian Nynorsk & nn & Indo-European & Germanic & Latin & \Checkmark & 486,055 \\
        Occitan & oc & Indo-European & Italic & Latin & \Checkmark & 35,791 \\
        Sinhala & si & Indo-European & Indo-Aryan & Sinhala & \XSolidBrush & 979,109 \\
        Tajik & tg & Indo-European & Iranian & Cyrillic & \XSolidBrush & 193,882 \\
        Turkmen & tk & Turkic & Common Turkic & Latin & \Checkmark & 13,110 \\
        Tatar & tt & Turkic & Common Turkic & Cyrillic & \XSolidBrush & 100,843 \\
        Uyghur & ug & Turkic & Common Turkic & Arabic & \Checkmark & 72,170 \\
        Northern Uzbek & uz & Turkic & Common Turkic & Latin & \Checkmark & 173,157 \\
        Eastern Yiddish & yi & Indo-European & Germanic & Hebrew & \XSolidBrush & 15,010 \\
        \midrule
        Total & & & & & & 4,498,632 \\
        \bottomrule
    \end{tabular}
    }
    \caption{\textbf{Statistics of training data for BLOOMZ+24}: 24 unseen, low-resource languages for BLOOMZ. \Checkmark and \XSolidBrush indicate whether script is seen or unseen.}
    \label{tab:opus-100}
\end{table*}

\begin{table*}[t]
    \centering
    \resizebox{\linewidth}{!}{
    \begin{tabular}{l|lrrr|lrrr}
        \toprule
        \multirow{2}{*}{ Languages } & \multicolumn{4}{c|}{ Zero-shot Directions } & \multicolumn{4}{c}{ Supervised Directions } \\
        & Directions & BLEU & chrF++ & COMET & Directions & BLEU & chrF++ & COMET \\
        \toprule
        \multirow{6}{*}{ de-nl-ru } & \multirow{2}{*}{overall} & \multirow{2}{*}{\textbf{8.94}} & \multirow{2}{*}{\textbf{23.53}} & \multirow{2}{*}{\textbf{60.67}} & en$\rightarrow$xx & 16.70 & 31.83 & 68.98 \\
        & & & & & xx$\rightarrow$en & \textbf{25.18} & 45.00 & \textbf{76.45} \\
        & seen$\rightarrow$seen & 14.00 & \textbf{27.58} & \textbf{70.59} & en$\rightarrow$seen & 15.97 & 28.53 & \textbf{72.69} \\
        & seen$\rightarrow$unseen & \textbf{6.49} & \textbf{23.01} & \textbf{54.92} & en$\rightarrow$unseen & \textbf{17.43} & \textbf{35.13} & \textbf{65.27} \\
        & unseen$\rightarrow$seen & \textbf{9.50} & \textbf{21.90} & \textbf{64.69} & seen$\rightarrow$en & 25.33 & 46.70 & 77.51 \\
        & unseen$\rightarrow$unseen & 6.73 & 22.70 & 53.34 & unseen$\rightarrow$en & \textbf{25.03} & \textbf{43.30} & \textbf{75.39} \\
        \bottomrule
        \multirow{6}{*}{ ar-de-fr-nl-ru-zh } & \multirow{2}{*}{overall} & \multirow{2}{*}{\textbf{12.07}} & \multirow{2}{*}{\textbf{26.67}} & \multirow{2}{*}{63.13} & en$\rightarrow$xx & \textbf{21.62} & \textbf{36.12} & \textbf{70.94} \\
        & & & & & xx$\rightarrow$en & \textbf{28.92} & \textbf{48.60} & \textbf{77.50} \\
        & seen$\rightarrow$seen & \textbf{23.52} & \textbf{36.13} & \textbf{76.62} & en$\rightarrow$seen & \textbf{26.87} & \textbf{38.40} & \textbf{78.40} \\
        & seen$\rightarrow$unseen & \textbf{7.16} & 24.48 & 55.02 & en$\rightarrow$unseen & \textbf{16.37} & \textbf{33.83} & 63.49 \\
        & unseen$\rightarrow$seen & \textbf{12.91} & \textbf{25.23} & \textbf{68.91} & seen$\rightarrow$en & \textbf{32.57} & \textbf{53.70} & \textbf{80.06} \\
        & unseen$\rightarrow$unseen & 6.73 & \textbf{22.65} & 53.12 & unseen$\rightarrow$en & \textbf{25.27} & \textbf{43.50} & \textbf{74.93} \\
        \bottomrule
    \end{tabular}
    }
    \caption{\textbf{Results of BLOOMZ+3 with MT+Align+Hint+Revise.} Co-referencing Tab.~\ref{tab:res5}, scores that surpass the MTInstruct baseline are marked in \textbf{bold}.}
    \label{tab:bloomz+3-app}
\end{table*}

\appendix

\section{Training Data Statistics}
\label{app:data}
Training data statistics of BLOOMZ+24 are shown in Tab.~\ref{tab:opus-100}. Several selected languages involved previously unseen scripts by BLOOMZ, but such fine-tuning is practical as BLOOMZ is a byte-level model with the potential to adapt to any language. Note that our proposed methods can be applied to any byte-level generative LLMs.

\section{Implementation Details}
\label{app:training}
We employed $128$ V100 GPUs for the BLOOMZ+24 and $32$ V100 GPUs for the BLOOMZ+3 experiments. The batch sizes were configured at $4$ sentences for BLOOMZ-7b1 and $8$ sentences for both BLOOMZ-3b and BLOOMZ-1b1, per GPU device. We configured LoRA with a rank of $8$, an alpha of $32$, and a dropout of $0.1$. Consequently, the BLOOMZ-7b1, BLOOMZ-3b, and BLOOMZ-1b1 models had $3.9$M, $2.5$M, and $1.2$M trainable parameters, respectively, constituting approximately $0.05$ - $0.10$\% of the parameters in the original models. We conducted training for $5$ epochs, ensuring a stable convergence is achieved. To facilitate this stability, we introduced a warm-up ratio of $0.03$ into our training process. Maximum input and output length were set as $384$. $S$ for HintInstruct was set as $5$ at most. Additionally, we used mixed precision training~\cite{DBLP:conf/iclr/MicikeviciusNAD18} to expedite computation using DeepSpeed~\cite{DBLP:conf/kdd/RasleyRRH20}. We tuned the optimal learning rate for each individual experiment according to validation loss. We conducted all experiments once due to computational resource constraints and reported the average scores across all languages.

\section{Results of MT+Align+Hint+Revise for BLOOMZ+3}
\label{app:zero}

We present the results in Tab.~\ref{tab:bloomz+3-app}. Co-referencing the results in Tab.~\ref{tab:res5}, compared with MT+Align, we observed a clear advantage for the MT+Align+Hint+Revise setting in supervised directions involving English (en$\rightarrow$seen and seen$\rightarrow$en) in the ar-fr-de-nl-ru-zh setting. This result suggested that AlignInstruct's variants played a crucial role in preserving the BLOOMZ's capabilities for supported languages. However, in all other scenarios, AlignInstruct alone proved sufficient to enhance the performance beyond the MTInstruct baseline, but hard to achieve further improvements with additional instructions.

\section{Representation Change of BLOOMZ+3}
\label{app:embedding}

\begin{figure}[t]
    \centering
    \includegraphics[width=\linewidth]{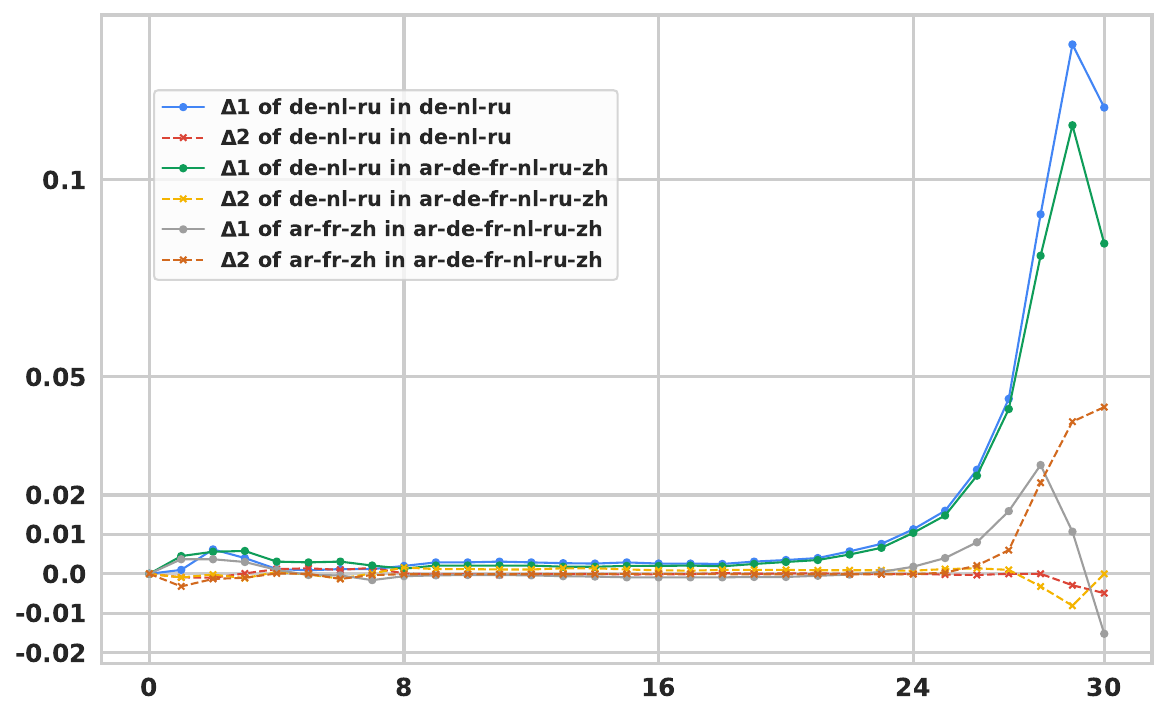}
    \caption{\textbf{Differences in cosine similarity of layer-wise embeddings for BLOOMZ+3.} $\Delta$1 represents the changes from the unmodified BLOOMZ to the one on MTInstruct, and $\Delta$2 from MTInstruct to MT+Align.}
    \label{fig:cosine2}
\end{figure}

The representation change observed in de-nl-ru was consistent with the findings presented in Sec.~\ref{sec:4.5}, which highlighted an initial increase in cross-lingual alignment in the early layers, followed by a decrease in the final layers. When mixing fine-tuning data with supported languages, the changes exhibited more intricate patterns. As illustrated by ar-fr-zh in ar-de-fr-nl-ru-zh in Fig.~\ref{fig:cosine2}, sentence alignment declined after MTInstruct fine-tuning but elevated after further combining with AlignInstruct. We leave the interpretation of this nuanced behavior in future work.

\section{Examples of HintInstruct and ReviseInstruct}
\label{app:hint-revise}

\begin{figure}[t]
    \centering
    \includegraphics[width=\linewidth]{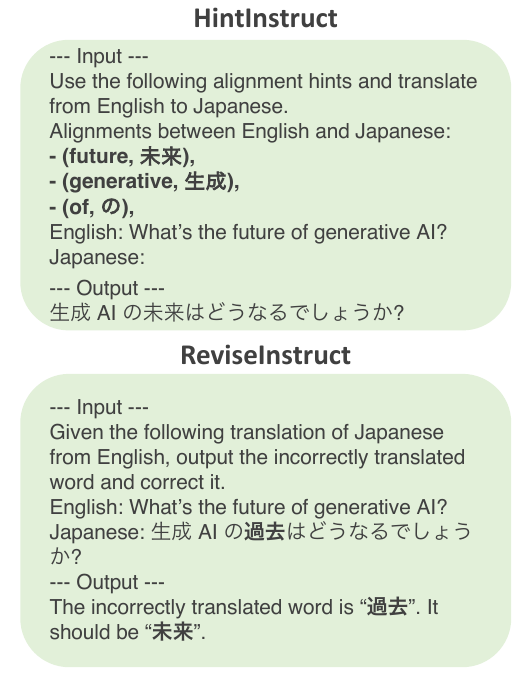}
    \caption{Examples of HintInstruct and ReviseInstruct.}
    \label{fig:hint-revise}
\end{figure}

We illustrated examples of HintInstruct and ReviseInstruct in Fig.~\ref{fig:hint-revise}.

\begin{table*}[t]
    \centering
    \resizebox{\linewidth}{!}{
    \begin{tabular}{l|rrr|rrr|rrr|rrr}
        \toprule
        \multirow{2}{*}{ Objective } & \multicolumn{3}{c|}{ OPUS en$\rightarrow$xx } & \multicolumn{3}{c|}{ OPUS xx$\rightarrow$en } & \multicolumn{3}{c|}{ Flores en$\rightarrow$xx } & \multicolumn{3}{c}{ Flores xx$\rightarrow$en } \\
        & BLEU & chrF++ & COMET & BLEU & chrF++ & COMET & BLEU & chrF++ & COMET & BLEU & chrF++ & COMET \\
        \toprule
        MTInstruct & 11.54 & 25.33 & 64.54 & 18.59 & 33.25 & 69.84 & 3.30 & 17.10 & 40.58 & 11.37 & 27.14 & 56.33 \\
        \hline
        MT+Mono-full & 9.89 & 22.42 & 62.52 & 15.43 & 29.04 & 66.64 & 3.00 & 16.68 & 40.49 & 10.26 & 25.15 & 54.17 \\
        MT+Mono-half & 10.23 & 22.45 & 62.22 & 15.51 & 29.65 & 67.29 & 3.18 & 16.91 & 40.57 & 10.66 & 26.15 & 54.80 \\
        MT+Mono-full+Align & 10.15 & 22.35 & 62.22 & 15.72 & 29.86 & 67.70 & 3.07 & 16.59 & \textbf{40.78} & 10.61 & 25.58 & 55.17 \\
        MT+Mono-half+Align & 10.09 & 22.61 & 62.98 & 16.00 & 30.34 & 67.96 & 3.10 & 16.75 & \textbf{40.70} & 10.79 & 26.27 & 55.40 \\
        MT+Mono-full+Align+Hint+Revise & 10.33 & 23.04 & 63.19 & 17.16 & 31.61 & 68.26 & 3.23 & 16.70 & \textbf{40.90} & 10.98 & 26.18 & 55.50 \\
        MT+Mono-half+Align+Hint+Revise & 10.62 & 23.10 & 62.92 & 17.32 & 31.80 & 68.56 & 3.20 & 16.93 & \textbf{41.00} & 11.09 & 26.77 & 55.99 \\
        \bottomrule
        \hline
    \end{tabular}
    }
    \caption{\textbf{Results of BLOOMZ+24 fine-tuned incorporating monolingual instructions on BLOOMZ-7b1.} Scores that surpass the MTInstruct baseline are marked in \textbf{bold}.} 
    \label{tab:res4}
\end{table*}

\section{Assessing Monolingual Instructions}
\label{app:mono}

New language capabilities may be induced through continual pre-training on monolingual next-word prediction tasks \cite{yong-etal-2023-bloom}. The coherence of the generated sentences is crucial in MT~\cite{wang-etal-2020-multi,liu-etal-2020-multilingual-denoising}, especially when the target languages are unseen and low-resource. We examined the significance of this approach in fostering the translation quality. We reused the same parallel corpora to avoid introducing additional monolingual datasets. 


Given a monolingual sentence, $\left(x_i\right)_1^N$, with length $N$ in an unseen language $X$. The LLM is incrementally trained on the following task:

\begin{itemize}
    \item \noindent\textbf{Input}: ``Given the context, complete the following sentence: $x_1 x_2 \ldots x_{l<N}$,''
    \item \noindent\textbf{Output}: ``$x_{l+1} x_{l+2} \ldots x_N$.''
\end{itemize}

We conducted experiments with two MonoInstruct settings: \textbf{MonoInstruct-full}, an objective to generate the entire sentence, and \textbf{MonoInstruct-half} for generating the latter half of the sentence given the first half, inspired by GPT~\cite{radford2018improving} and MASS~\cite{DBLP:conf/icml/SongTQLL19}, respectively. We reported the MonoInstruct results in Tab.~\ref{tab:res4}. Firstly, we observed that fine-tuning MTInstruct in conjunction with either MonoInstruct-full or MonoInstruct-half harms the MT performance, which could be attributed to the inherent difficulty of monolingual instruction tasks and the limited amount of monolingual data. We found that the simpler MT+Mono-half yielded better results than MT+Mono-full as richer contexts were provided. However, MonoInstruct still did not improve the MTInstruct baseline. Secondly, further combining MonoInstrcut with AlignInstruct variants yielded improvements compared with MT+Mono-full (or half), but underperformed the MTInstruct baseline. This suggested that improving MT performance with monolingual instructions is challenging without access to additional monolingual data.

\section{Inference using Different MT Prompts}
\label{app:prompts}

We investigated the performance of fine-tuned models when using various MT prompts during inference, aiming to understand models' generalization capabilities with different test prompts. We examined five MT prompts for the fine-tuned models of BLOOMZ-7b1, following~\citet{DBLP:conf/icml/0006HB23}, which are presented in Tab.~\ref{tab:mt-prompt}. The results, showcased in Tab.~\ref{tab:prompt}, revealed that in comparison to the default prompt used during fine-tuning, the translation performance tended to decline when using other MT prompts. We observed that MT+Align consistently surpasses MTInstruct for xx$\rightarrow$en translations, though the results were mixed for en$\rightarrow$xx directions. Certain prompts, such as PROMPT-3 and PROMPT-4, exhibited a minor performance drop, while others significantly impacted translation quality. These findings underscored the need for enhancing the models' ability to generalize across diverse MT prompts, potentially by incorporating a range of MT prompt templates during the fine-tuning process, as stated in the Limitations section.

\section{Zero-shot Translation using English as Pivot}
\label{app:pivot}

\begin{table}[t!]
    \centering
    \resizebox{0.85\linewidth}{!}{
    \begin{tabular}{ll}
        \toprule
        Prompt & Definition \\
        \toprule
        \multirow{3}{*}{PROMPT-default} & Translate from $Y$ to $X$. \\
        & $Y$: $y_1 y_2 \ldots y_M$. \\
        & $X$: \\
        \hline
        \multirow{2}{*}{PROMPT-1} & $Y$: $y_1 y_2 \ldots y_M$. \\
        & $X$: \\
        \hline
        \multirow{2}{*}{PROMPT-2} & $y_1 y_2 \ldots y_M$. \\
        & $X$: \\
        \hline
        \multirow{3}{*}{PROMPT-3} & Translate to $X$. \\
        & $Y$: $y_1 y_2 \ldots y_M$. \\
        & $X$: \\
        \hline
        \multirow{3}{*}{PROMPT-4} & Translate from $Y$ to $X$. \\
        & $y_1 y_2 \ldots y_M$. \\
        & $X$: \\
        \hline
        \multirow{3}{*}{PROMPT-5} & Translate to $X$. \\
        & $y_1 y_2 \ldots y_M$. \\
        & $X$: \\
        \bottomrule
    \end{tabular}
    }
    \caption{\textbf{MT prompt variants investigated for fine-tuned models.} These MT prompts are following the design in~\citet{DBLP:conf/icml/0006HB23}.}
    \label{tab:mt-prompt}
\end{table}

\begin{table*}[t!]
    \centering
    \resizebox{0.8\linewidth}{!}{
    \begin{tabular}{ll|rrr|rrr}
        \toprule
        \multirow{2}{*}{Prompt} & \multirow{2}{*}{Objective} & \multicolumn{3}{c|}{en$\rightarrow$xx} & \multicolumn{3}{c}{xx$\rightarrow$en} \\
        & & BLEU & chrF++ & COMET & BLEU & chrF++ & COMET \\
        \toprule
        \multirow{2}{*}{PROMPT-default} & MTInstruct & 11.54 & 25.33 & 64.54 & 18.59 & 33.25 & 69.84 \\
        & MT+Align & \textbf{12.28} & \textbf{26.17} & \textbf{65.54} & \textbf{18.72} & \textbf{34.02} & \textbf{70.69} \\
        \hline 
        \multirow{2}{*}{PROMPT-1} & MTInstruct & 5.29 & 11.31 & 50.20 & 7.87 & 20.08 & 57.46 \\
        & MT+Align & \textbf{5.30} & \textbf{11.38} & \textbf{50.95} & \textbf{8.93} & \textbf{20.77} & \textbf{58.38} \\
        \hline
        \multirow{2}{*}{PROMPT-2} & MTInstruct & 2.20 & 6.68 & 45.56 & 7.15 & 19.08 & 57.22 \\
        & MT+Align & 1.91 & 5.35 & 43.84 & \textbf{7.61} & 18.80 & 56.76 \\
        \hline
        \multirow{2}{*}{PROMPT-3} & MTInstruct & 10.59 & 22.69 & 62.65 & 15.85 & 29.93 & 67.59 \\
        & MT+Align & 9.20 & 20.80 & 60.96 & \textbf{16.17} & \textbf{30.58} & \textbf{68.70} \\
        \hline
        \multirow{2}{*}{PROMPT-4} & MTInstruct & 8.67 & 20.73 & 61.50 & 15.20 & 28.95 & 66.61 \\
        & MT+Align & \textbf{8.91} & 20.53 & \textbf{61.64} & \textbf{16.25} & \textbf{30.67} & \textbf{67.94} \\
        \hline
        \multirow{2}{*}{PROMPT-5} & MTInstruct & 6.61 & 14.55 & 55.99 & 10.88 & 22.41 & 61.40 \\
        & MT+Align & 6.02 & 12.28 & 52.42 & \textbf{11.83} & \textbf{23.85} & \textbf{62.09} \\
        \bottomrule
    \end{tabular}
    }
    \caption{\textbf{Results of using different MT prompts for BLOOMZ-7b1 fine-tuned models during inference.} Refer to Tab.~\ref{tab:mt-prompt} for details about definitions of different MT prompts. We report the average results for the BLOOMZ+24 setting. Results better than the MTInstruct baseline are marked in \textbf{bold}.}
    \label{tab:prompt}
\end{table*}

\begin{table*}[t!]
    \centering
    \resizebox{\linewidth}{!}{
    \begin{tabular}{llll|llll}
        \toprule
        \textbf{MTInstruct} & BLEU & chrF++ & COMET & \textbf{MT+Align} & BLEU & chrF++ & COMET \\
        \toprule
        overall & 11.79 & 26.36 & 63.22 & overall & \textbf{12.13} & \textbf{26.65} & \textbf{63.23} \\
        \hline
        seen$\rightarrow$seen & 22.68 & 35.32 & 76.39 & seen$\rightarrow$seen & \textbf{23.67} & \textbf{36.53} & \textbf{76.89} \\
        seen$\rightarrow$unseen & 7.10 & 24.50 & 55.18 & seen$\rightarrow$unseen & \textbf{7.27} & 24.32 & 54.96 \\
        unseen$\rightarrow$seen & 12.56 & 24.74 & 68.83 & unseen$\rightarrow$seen & \textbf{12.92} & \textbf{25.29} & \textbf{69.10} \\
        unseen$\rightarrow$unseen & 6.78 & 22.62 & 53.69 & unseen$\rightarrow$unseen & 6.68 & 22.30 & 53.19 \\
        \toprule
        \toprule
        \textbf{MTInstruct with English pivot} & BLEU & chrF++ & COMET & \textbf{MT+Align with English pivot} & BLEU & chrF++ & COMET \\
        \toprule
        overall & 12.99 & 28.01 & 65.38 & overall & \textbf{13.25} & \textbf{28.30} & \textbf{65.57} \\
        \hline
        seen$\rightarrow$seen & 23.10 & 35.30 & 76.30 & seen$\rightarrow$seen & \textbf{23.48} & \textbf{35.57} & \textbf{76.43} \\
        seen$\rightarrow$unseen & 9.00 & 27.67 & 59.54 & seen$\rightarrow$unseen & \textbf{9.28} & \textbf{28.03} & \textbf{59.73} \\
        unseen$\rightarrow$seen & 13.18 & 24.98 & 68.77 & unseen$\rightarrow$seen & \textbf{13.36} & \textbf{25.22} & \textbf{68.94} \\
        unseen$\rightarrow$unseen & 8.57 & 25.77 & 58.17 & unseen$\rightarrow$unseen & \textbf{8.83} & \textbf{26.07} & \textbf{58.42} \\
        \bottomrule
    \end{tabular}
    }
    \caption{\textbf{Results of BLOOMZ+3 using English as a pivot language for zero-shot translation evaluation.} Results of MT+Align surpassing corresponding those of MTInstruct are marked in \textbf{bold}.}
    \label{tab:pivot}
\end{table*}

Pivot translation serves as a robust technique for zero-shot translation, especially given that we used English-centric data during fine-tuning. In Tab.~\ref{tab:pivot}, we present results that utilize English as an intermediary pivot for translations between non-English language pairs. Our findings indicated that employing the English pivot typically yielded an enhancement of approximately 1.1 - 1.2 BLEU points compared to direct translations in zero-shot directions when fine-tuning BLOOMZ. When contrasting the MTInstruct baseline with our proposed MT+Align, we observed that combining AlignInstruct consistently boosted performance in pivot translation scenarios.

\section{Per Language Result Details of BLOOMZ+24 and BLOOMZ+3}
We present per language detailed results of original BLOOMZ-7b1 and fine-tuned BLOOMZ-7b1 models in Tab.~\ref{7b1},~\ref{7b1-A},~\ref{7b1-AD},~\ref{7b1-6langs},~\ref{7b1-3langs-A},~\ref{7b1-3langs-AD},~\ref{7b1-6langs-A},~\ref{7b1-6langs-AD}, respectively for the BLOOMZ+24 and BLOOMZ+3 settings.

\begin{table*}[t]
\centering
\resizebox{\linewidth}{!}{
\begin{tabular}{l|rrr|rrr|rrr|rrr}
\toprule
\multirow{2}{*}{Language} & \multicolumn{3}{c|}{OPUS en$\rightarrow$xx} & \multicolumn{3}{c|}{OPUS xx$\rightarrow$en} & \multicolumn{3}{c|}{Flores en$\rightarrow$xx} & \multicolumn{3}{c}{Flores xx$\rightarrow$en} \\
& BLEU & chrF++ & COMET & BLEU & chrF++ & COMET & BLEU & chrF++ & COMET & BLEU & chrF++ & COMET \\
\toprule
af & 3.8 & 13.2 & 56.38 & 7.6 & 22.0 & 59.14 & 2.6 & 14.9 & 33.60 & 20.1 & 38.0 & 65.61 \\
am & 0.1 & 0.3 & 33.17 & 0.5 & 8.3 & 43.57 & 0.3 & 0.6 & 30.65 & 1.9 & 12.6 & 46.24 \\
be & 4.2 & 5.1 & 47.26 & 7.3 & 17.5 & 48.57 & 0.4 & 3.3 & 31.58 & 4.2 & 22.3 & 49.27 \\
cy & 2.7 & 10.5 & 53.21 & 6.2 & 16.0 & 53.25 & 1.2 & 11.2 & 34.17 & 6.0 & 20.3 & 53.45 \\
ga & 1.2 & 10.6 & 42.85 & 4.0 & 16.4 & 46.05 & 1.2 & 11.6 & 33.94 & 5.5 & 19.6 & 46.97 \\
gd & 9.3 & 16.0 & 51.40 & 47.6 & 55.9 & 59.30 & 1.2 & 11.2 & 36.28 & 4.2 & 18.8 & 43.73 \\
gl & 4.5 & 25.6 & 64.93 & 17.2 & 36.7 & 66.07 & 13.4 & 38.5 & 74.77 & 51.0 & 67.8 & 85.77 \\
ha & 0.1 & 5.4 & 38.42 & 0.3 & 11.2 & 42.58 & 1.5 & 10.2 & 35.77 & 6.9 & 18.9 & 47.37 \\
ka & 0.3 & 1.9 & 31.97 & 0.6 & 9.2 & 44.48 & 0.4 & 1.4 & 28.81 & 2.4 & 17.0 & 47.57 \\
kk & 4.3 & 4.9 & 50.51 & 5.1 & 14.2 & 51.51 & 0.5 & 1.6 & 33.66 & 5.1 & 19.8 & 51.40 \\
km & 2.8 & 4.5 & 51.68 & 3.9 & 11.1 & 50.40 & 0.8 & 2.9 & 39.56 & 5.6 & 16.2 & 50.42 \\
ky & 10.0 & 10.6 & 54.23 & 10.3 & 24.0 & 55.99 & 0.6 & 1.6 & 30.19 & 3.8 & 17.9 & 48.05 \\
li & 6.6 & 16.2 & - & 5.9 & 24.8 & - & 2.0 & 14.9 & - & 9.8 & 29.8 & - \\
my & 1.8 & 2.4 & 45.44 & 3.0 & 5.0 & 48.33 & 0.4 & 0.8 & 29.58 & 1.0 & 3.7 & 44.15 \\
nb & 5.8 & 18.2 & 57.01 & 13.9 & 33.0 & 56.37 & 3.9 & 19.3 & 46.74 & 19.8 & 40.3 & 63.56 \\
nn & 6.3 & 18.6 & 62.33 & 8.9 & 25.3 & 56.28 & 3.7 & 19.7 & 41.75 & 16.9 & 37.5 & 62.37 \\
oc & 6.0 & 13.6 & - & 5.1 & 18.6 & - & 9.6 & 33.6 & - & 53.0 & 68.5 & - \\
si & 0.6 & 2.0 & 41.84 & 1.6 & 9.4 & 48.58 & 0.5 & 1.4 & 28.08 & 1.6 & 9.1 & 42.67 \\
tg & 0.4 & 1.4 & - & 1.1 & 11.8 & - & 0.4 & 1.5 & - & 3.3 & 18.0 & - \\
tk & 7.9 & 10.6 & - & 5.3 & 13.0 & - & 0.7 & 8.7 & - & 4.2 & 20.1 & - \\
tt & 0.0 & 1.0 & - & 0.2 & 13.3 & - & 0.3 & 1.4 & - & 4.2 & 20.2 & - \\
ug & 0.0 & 0.4 & 32.44 & 0.3 & 11.2 & 45.69 & 0.3 & 0.9 & 31.34 & 3.0 & 16.5 & 48.99 \\
uz & 0.7 & 2.1 & 35.94 & 1.0 & 12.8 & 41.86 & 1.5 & 11.5 & 40.65 & 3.1 & 18.7 & 49.43 \\
yi & 7.3 & 16.5 & 57.47 & 4.0 & 23.0 & 63.91 & 0.7 & 1.7 & 33.22 & 2.1 & 15.6 & 41.87 \\
\hline
avg. & 3.61 & 8.82 & 47.81 & 6.70 & 18.49 & 51.68 & 2.00 & 9.35 & 36.54 & 9.95 & 24.47 & 52.05 \\
\bottomrule
\end{tabular}
}
\caption{Detailed results of BLOOMZ-7b1 without fine-tuning.}
\label{7b1}
\end{table*}

\begin{table*}[t]
\centering
\resizebox{\linewidth}{!}{
\begin{tabular}{l|rrr|rrr|rrr|rrr}
\toprule
\multirow{2}{*}{Language} & \multicolumn{3}{c|}{OPUS en$\rightarrow$xx} & \multicolumn{3}{c|}{OPUS xx$\rightarrow$en} & \multicolumn{3}{c|}{Flores en$\rightarrow$xx} & \multicolumn{3}{c}{Flores xx$\rightarrow$en} \\
& BLEU & chrF++ & COMET & BLEU & chrF++ & COMET & BLEU & chrF++ & COMET & BLEU & chrF++ & COMET \\
\toprule
af & 25.0 & 41.4 & 71.05 & 38.5 & 52.3 & 78.94 & 10.1 & 31.0 & 45.42 & 33.9 & 51.1 & 72.66 \\
am & 3.0 & 12.8 & 59.55 & 3.4 & 19.8 & 59.71 & 0.2 & 5.2 & 42.97 & 1.4 & 16.0 & 49.47 \\
be & 8.9 & 14.9 & 55.16 & 14.0 & 24.9 & 62.37 & 0.7 & 12.3 & 30.90 & 3.7 & 21.0 & 49.99 \\
cy & 20.2 & 38.0 & 71.55 & 33.2 & 49.3 & 77.72 & 5.0 & 20.3 & 38.38 & 13.1 & 30.2 & 57.47 \\
ga & 15.6 & 37.1 & 63.87 & 29.2 & 49.1 & 75.94 & 3.7 & 21.2 & 39.17 & 12.5 & 30.3 & 57.53 \\
gd & 13.1 & 24.7 & 62.14 & 66.0 & 69.6 & 77.70 & 2.2 & 19.6 & 40.75 & 7.1 & 22.3 & 50.05 \\
gl & 16.9 & 37.6 & 70.62 & 24.7 & 43.6 & 75.62 & 21.9 & 45.2 & 77.26 & 46.6 & 64.5 & 86.86 \\
ha & 12.3 & 32.7 & 71.75 & 10.0 & 29.8 & 64.51 & 1.9 & 17.1 & 49.24 & 6.8 & 22.1 & 48.81 \\
ka & 4.6 & 18.1 & 67.39 & 10.0 & 24.3 & 60.50 & 0.3 & 6.8 & 27.46 & 1.5 & 14.9 & 46.10 \\
kk & 12.6 & 19.5 & 66.07 & 14.6 & 28.2 & 71.80 & 0.8 & 13.0 & 35.76 & 3.9 & 19.7 & 52.24 \\
km & 19.7 & 25.2 & 63.24 & 13.9 & 32.1 & 75.02 & 0.5 & 12.3 & 35.60 & 6.2 & 22.4 & 56.45 \\
ky & 16.0 & 20.5 & 66.27 & 21.1 & 33.8 & 73.06 & 0.9 & 12.7 & 36.10 & 3.0 & 17.5 & 50.40 \\
li & 13.5 & 32.8 & - & 21.3 & 35.7 & - & 3.3 & 19.9 & - & 14.6 & 31.4 & - \\
my & 6.2 & 14.3 & 58.04 & 5.2 & 15.6 & 63.65 & 0.2 & 12.9 & 40.37 & 1.3 & 12.7 & 48.38 \\
nb & 12.7 & 30.4 & 63.27 & 22.2 & 42.1 & 76.74 & 7.9 & 28.4 & 44.15 & 25.6 & 44.3 & 72.56 \\
nn & 18.3 & 38.0 & 77.18 & 27.1 & 47.7 & 81.80 & 7.3 & 25.7 & 45.35 & 24.3 & 42.9 & 70.06 \\
oc & 10.0 & 20.0 & - & 13.4 & 27.1 & - & 8.0 & 27.5 & - & 46.9 & 63.5 & - \\
si & 5.2 & 21.4 & 68.16 & 11.5 & 26.4 & 70.79 & 0.9 & 12.9 & 41.73 & 3.7 & 19.2 & 57.41 \\
tg & 5.5 & 22.0 & - & 8.0 & 25.9 & - & 1.1 & 15.8 & - & 3.1 & 19.6 & - \\
tk & 24.4 & 26.7 & - & 30.4 & 37.8 & - & 0.7 & 10.8 & - & 3.9 & 18.8 & - \\
tt & 1.9 & 17.6 & - & 3.6 & 19.6 & - & 0.4 & 13.7 & - & 1.6 & 14.3 & - \\
ug & 1.2 & 19.7 & 49.76 & 4.2 & 21.2 & 61.34 & 0.4 & 12.9 & 35.88 & 1.7 & 16.7 & 50.29 \\
uz & 3.1 & 18.2 & 62.12 & 5.7 & 22.0 & 61.12 & 0.5 & 3.6 & 34.67 & 3.9 & 18.8 & 50.32 \\
yi & 7.1 & 24.3 & 59.13 & 14.9 & 20.2 & 58.66 & 0.3 & 9.5 & 29.77 & 2.5 & 17.2 & 43.27 \\
\hline
avg. & 11.54 & 25.33 & 64.54 & 18.6 & 33.25 & 68.84 & 3.30 & 17.10 & 40.58 & 11.37 & 27.14 & 56.33 \\
\bottomrule
\end{tabular}
}
\caption{Detailed results of BLOOMZ-7b1 fine-tuned with MTInstruct for BLOOMZ+24.}
\label{7b1-A}
\end{table*}

\begin{table*}[t]
\centering
\resizebox{\linewidth}{!}{
\begin{tabular}{l|rrr|rrr|rrr|rrr}
\toprule
\multirow{2}{*}{Language} & \multicolumn{3}{c|}{OPUS en$\rightarrow$xx} & \multicolumn{3}{c|}{OPUS xx$\rightarrow$en} & \multicolumn{3}{c|}{Flores en$\rightarrow$xx} & \multicolumn{3}{c}{Flores xx$\rightarrow$en} \\
& BLEU & chrF++ & COMET & BLEU & chrF++ & COMET & BLEU & chrF++ & COMET & BLEU & chrF++ & COMET \\
\toprule
af & 25.0 & 41.9 & 70.72 & 36.9 & 52.2 & 78.68 & 10.6 & 31.9 & 45.84 & 33.5 & 51.1 & 72.84 \\
am & 3.4 & 13.2 & 60.62 & 4.9 & 22.8 & 62.43 & 0.3 & 5.4 & 44.20 & 1.4 & 16.4 & 51.05 \\
be & 8.3 & 14.5 & 55.23 & 13.9 & 25.1 & 62.72 & 0.8 & 12.5 & 30.93 & 3.6 & 20.6 & 49.14 \\
cy & 20.6 & 39.0 & 71.73 & 33.8 & 49.4 & 77.55 & 4.7 & 20.3 & 38.70 & 14.6 & 31.5 & 58.34 \\
ga & 17.6 & 39.3 & 65.76 & 32.6 & 52.7 & 77.49 & 3.4 & 21.4 & 39.99 & 13.6 & 31.6 & 58.73 \\
gd & 15.6 & 27.2 & 62.09 & 48.1 & 55.4 & 75.90 & 2.3 & 20.3 & 40.81 & 7.4 & 22.0 & 49.99 \\
gl & 17.1 & 37.2 & 70.85 & 24.4 & 43.3 & 75.90 & 21.7 & 44.9 & 77.09 & 45.6 & 63.5 & 86.60 \\
ha & 14.6 & 35.0 & 73.34 & 11.4 & 31.3 & 65.69 & 1.9 & 17.3 & 50.88 & 7.4 & 22.5 & 49.57 \\
ka & 4.9 & 18.9 & 67.54 & 10.5 & 25.3 & 61.27 & 0.3 & 6.9 & 27.61 & 2.1 & 16.0 & 47.04 \\
kk & 12.3 & 19.3 & 65.73 & 15.6 & 28.0 & 71.01 & 0.9 & 13.0 & 35.86 & 4.1 & 19.8 & 52.43 \\
km & 20.4 & 26.5 & 63.38 & 14.4 & 35.2 & 75.62 & 0.6 & 12.5 & 35.44 & 7.1 & 22.9 & 57.81 \\
ky & 15.8 & 19.6 & 64.74 & 23.3 & 35.8 & 74.70 & 0.9 & 13.3 & 36.71 & 2.9 & 17.4 & 50.06 \\
li & 13.2 & 29.4 & - & 22.3 & 38.2 & - & 3.1 & 19.7 & - & 12.5 & 28.7 & - \\
my & 7.6 & 15.4 & 58.84 & 6.3 & 18.0 & 66.45 & 0.3 & 13.3 & 40.97 & 1.2 & 14.4 & 50.79 \\
nb & 13.5 & 31.4 & 64.08 & 24.2 & 44.2 & 77.58 & 7.9 & 28.7 & 44.12 & 25.5 & 44.9 & 72.72 \\
nn & 19.0 & 38.0 & 77.61 & 28.5 & 47.7 & 81.68 & 7.0 & 26.7 & 46.14 & 25.8 & 44.1 & 70.55 \\
oc & 9.1 & 19.3 & - & 13.5 & 27.5 & - & 7.5 & 25.9 & - & 47.3 & 63.8 & - \\
si & 5.1 & 22.1 & 69.60 & 13.9 & 29.1 & 72.51 & 1.1 & 13.1 & 43.01 & 5.6 & 22.7 & 61.89 \\
tg & 6.6 & 23.7 & - & 8.8 & 27.2 & - & 0.9 & 15.6 & - & 3.4 & 19.9 & - \\
tk & 27.2 & 26.2 & - & 31.2 & 38.7 & - & 0.7 & 11.4 & - & 3.8 & 18.2 & - \\
tt & 2.1 & 18.6 & - & 5.0 & 21.5 & - & 0.4 & 13.3 & - & 1.5 & 13.7 & - \\
ug & 1.1 & 20.7 & 51.12 & 5.5 & 23.4 & 63.42 & 0.4 & 13.8 & 37.51 & 2.1 & 16.3 & 50.45 \\
uz & 3.5 & 18.6 & 62.09 & 7.4 & 23.3 & 62.01 & 0.2 & 1.9 & 34.50 & 3.7 & 18.2 & 50.09 \\
yi & 11.1 & 33.1 & 70.13 & 12.8 & 21.2 & 60.47 & 0.4 & 9.8 & 30.08 & 2.6 & 17.0 & 42.57 \\
\hline
avg. & 12.28 & 26.17 & 65.54 & 18.72 & 34.02 & 70.69 & 3.26 & 17.20 & 41.07 & 11.60 & 27.38 & 56.98 \\
\bottomrule
\end{tabular}
}
\caption{Detailed results of BLOOMZ-7b1 fine-tuned with MT+Align for BLOOMZ+24.}
\label{7b1-AD}
\end{table*}

\begin{table*}
\centering
\resizebox{\linewidth}{!}{
\begin{tabular}{lrrrc|lrrr}
\toprule
Zero-shot & BLEU & chrF++ & COMET &  & Supervised & BLEU & chrF++ & COMET \\
\toprule
ar-de & 1.4 & 14.8 & 56.19 &  & en-ar & 11.1 & 32.4 & 75.66 \\
ar-fr & 21.9 & 46.1 & 74.19 &  & en-de & 12.2 & 29.2 & 59.16 \\
ar-nl & 0.6 & 11.2 & 56.59 &  & en-fr & 26.8 & 49.2 & 77.42 \\
ar-ru & 3.1 & 6.2 & 48.41 &  & en-nl & 2.0 & 16.0 & 46.52 \\
ar-zh & 18.4 & 14.4 & 73.65 &  & en-ru & 5.7 & 16.1 & 49.00 \\
de-ar & 2.0 & 17.8 & 64.91 &  & en-zh & 22.5 & 17.0 & 77.90 \\
de-fr & 12.0 & 33.4 & 63.45 &  & avg. & 13.38 & 26.65 & 64.28 \\
de-nl & 3.7 & 17.9 & 47.30 &  &  &  &  &  \\
de-ru & 1.3 & 11.8 & 45.53 &  &  &  &  &  \\
de-zh & 8.9 & 7.6 & 61.52 &  &  &  &  &  \\
fr-ar & 11.2 & 33.4 & 74.20 &  &  & BLEU & chrF++ & COMET \\
fr-de & 4.6 & 23.4 & 48.83 &  & ar-en & 26.7 & 48.4 & 78.12 \\
fr-nl & 2.8 & 17.2 & 52.14 &  & de-en & 21.1 & 38.5 & 71.99 \\
fr-ru & 3.1 & 10.4 & 45.12 &  & fr-en & 27.7 & 49.8 & 79.46 \\
fr-zh & 20.9 & 17.0 & 76.20 &  & nl-en & 12.3 & 31.1 & 61.29 \\
nl-ar & 1.3 & 13.2 & 59.46 &  & ru-en & 17.9 & 36.6 & 68.40 \\
nl-de & 5.9 & 22.8 & 46.49 &  & zh-en & 24.5 & 47.9 & 77.08 \\
nl-fr & 9.6 & 29.6 & 58.30 &  & avg. & 21.70 & 42.05 & 72.72 \\
nl-ru & 0.8 & 9.0 & 42.83 &  &  &  &  &  \\
nl-zh & 3.3 & 3.7 & 53.96 &  &  &  &  &  \\
ru-ar & 6.5 & 25.3 & 68.38 &  &  &  &  &  \\
ru-de & 2.0 & 17.0 & 48.06 &  &  &  &  &  \\
ru-fr & 15.7 & 38.7 & 67.54 &  &  &  &  &  \\
ru-nl & 0.5 & 10.5 & 46.14 &  &  &  &  &  \\
ru-zh & 10.7 & 11.3 & 67.18 &  &  &  &  &  \\
zh-ar & 8.6 & 29.7 & 73.47 &  &  &  &  &  \\
zh-de & 1.6 & 17.6 & 49.90 &  &  &  &  &  \\
zh-fr & 20.7 & 44.1 & 75.79 &  &  &  &  &  \\
zh-nl & 0.6 & 10.4 & 48.53 &  &  &  &  &  \\
zh-ru & 2.9 & 8.6 & 44.13 &  &  &  &  &  \\
\hline
avg. & 6.89 & 19.14 & 57.95 &  &  &  &  &  \\
seen$\rightarrow$seen & 16.95 & 30.78 & 74.58 &  & en$\rightarrow$seen & 20.13 & 32.87 & 76.99 \\
seen$\rightarrow$unseen & 2.30 & 13.31 & 49.98 &  & en$\rightarrow$unseen & 6.63 & 20.43 & 51.56 \\
unseen$\rightarrow$seen & 7.78 & 20.07 & 62.74 &  & seen$\rightarrow$en & 26.30 & 48.70 & 78.22 \\
unseen$\rightarrow$unseen & 2.37 & 14.83 & 46.06 &  & unseen$\rightarrow$en & 17.10 & 35.40 & 67.23 \\
\bottomrule

\end{tabular}
}
\caption{Detailed results of BLOOMZ-7b1 without fine-tuning.}
\label{7b1-6langs}
\end{table*}

\begin{table*}
\centering
\resizebox{\linewidth}{!}{
\begin{tabular}{lrrrc|lrrr}
\toprule
Zero-shot & BLEU & chrF++ & COMET &  & Supervised & BLEU & chrF++ & COMET \\
\toprule
ar-de & 4.7 & 20.9 & 56.43 &  & en-ar & 9.1 & 27.2 & 71.47 \\
ar-fr & 20.8 & 42.5 & 71.47 &  & en-de & 19.8 & 36.1 & 66.53 \\
ar-nl & 7.2 & 22.9 & 58.29 &  & en-fr & 23.0 & 44.5 & 74.98 \\
ar-ru & 5.0 & 21.0 & 54.73 &  & en-nl & 15.5 & 36.1 & 64.76 \\
ar-zh & 14.0 & 12.4 & 67.94 &  & en-ru & 14.2 & 30.3 & 62.82 \\
de-ar & 2.4 & 16.2 & 64.53 &  & en-zh & 20.7 & 17.9 & 74.97 \\
de-fr & 11.9 & 31.2 & 64.44 &  & avg. & 17.05 & 32.02 & 69.26 \\
de-nl & 9.4 & 28.1 & 54.22 &  &  &  &  &  \\
de-ru & 5.1 & 19.6 & 55.41 &  &  &  &  &  \\
de-zh & 4.2 & 5.8 & 55.26 &  &  &  &  &  \\
fr-ar & 10.1 & 29.1 & 70.72 &  &  & BLEU & chrF++ & COMET \\
fr-de & 8.6 & 27.7 & 53.77 &  & ar-en & 26.5 & 46.9 & 76.92 \\
fr-nl & 10.3 & 30.1 & 57.55 &  & de-en & 27.0 & 44.0 & 76.97 \\
fr-ru & 7.9 & 26.0 & 56.82 &  & fr-en & 27.5 & 49.0 & 78.80 \\
fr-zh & 18.1 & 18.5 & 72.24 &  & nl-en & 21.8 & 41.3 & 73.99 \\
nl-ar & 2.0 & 15.1 & 63.73 &  & ru-en & 24.8 & 43.6 & 74.23 \\
nl-de & 9.7 & 28.1 & 52.58 &  & zh-en & 23.2 & 45.3 & 76.83 \\
nl-fr & 13.2 & 32.3 & 65.17 &  & avg. & 25.13 & 45.02 & 76.29 \\
nl-ru & 5.1 & 18.6 & 55.13 &  &  &  &  &  \\
nl-zh & 3.0 & 5.4 & 54.34 &  &  &  &  &  \\
ru-ar & 5.9 & 15.0 & 60.36 &  &  &  &  &  \\
ru-de & 5.6 & 23.8 & 52.66 &  &  &  &  &  \\
ru-fr & 17.9 & 38.4 & 68.66 &  &  &  &  &  \\
ru-nl & 6.2 & 22.5 & 54.41 &  &  &  &  &  \\
ru-zh & 7.5 & 13.6 & 61.40 &  &  &  &  &  \\
zh-ar & 6.7 & 22.1 & 67.48 &  &  &  &  &  \\
zh-de & 3.3 & 19.6 & 51.75 &  &  &  &  &  \\
zh-fr & 17.4 & 38.9 & 73.00 &  &  &  &  &  \\
zh-nl & 4.8 & 19.3 & 54.41 &  &  &  &  &  \\
zh-ru & 3.5 & 17.9 & 49.02 &  &  &  &  &  \\
\hline
avg. & 8.38 & 22.75 & 59.93 &  &  &  &  &  \\
seen$\rightarrow$seen & 14.52 & 27.25 & 70.48 &  & en$\rightarrow$seen & 17.60 & 29.87 & 73.81 \\
seen$\rightarrow$unseen & 6.14 & 22.82 & 54.75 &  & en$\rightarrow$unseen & 16.50 & 34.17 & 64.70 \\
unseen$\rightarrow$seen & 7.56 & 19.22 & 61.99 &  & seen$\rightarrow$en & 25.73 & 47.07 & 77.52 \\
unseen$\rightarrow$unseen & 6.85 & 23.45 & 54.07 &  & unseen$\rightarrow$en & 24.53 & 42.97 & 75.06 \\
\bottomrule
\end{tabular}
}
\caption{Detailed results of BLOOMZ-7b1 fine-tuned with MTInstruct for BLOOMZ+3 de-nl-ru.}
\label{7b1-3langs-A}
\end{table*}

\begin{table*}
\centering
\resizebox{\linewidth}{!}{
\begin{tabular}{lrrrc|lrrr}
\toprule
Zero-shot & BLEU & chrF++ & COMET &  & Supervised & BLEU & chrF++ & COMET \\
\toprule
ar-de & 5.1 & 20.8 & 55.25 &  & en-ar & 8.4 & 26.0 & 70.45 \\
ar-fr & 20.3 & 42.5 & 71.78 &  & en-de & 21.1 & 36.7 & 67.15 \\
ar-nl & 6.4 & 21.6 & 57.48 &  & en-fr & 22.9 & 44.4 & 74.67 \\
ar-ru & 5.2 & 21.5 & 55.51 &  & en-nl & 16.1 & 36.8 & 65.26 \\
ar-zh & 16.0 & 14.1 & 69.55 &  & en-ru & 15.2 & 31.5 & 63.30 \\
de-ar & 2.4 & 16.3 & 64.01 &  & en-zh & 16.1 & 15.0 & 71.93 \\
de-fr & 13.5 & 34.3 & 66.25 &  & avg. & 16.63 & 31.73 & 68.79 \\
de-nl & 9.7 & 28.0 & 55.00 &  &  &  &  &  \\
de-ru & 5.3 & 19.6 & 55.61 &  &  &  &  &  \\
de-zh & 7.2 & 7.3 & 60.64 &  &  &  &  &  \\
fr-ar & 10.0 & 28.2 & 69.86 &  &  & BLEU & chrF++ & COMET \\
fr-de & 9.2 & 27.8 & 54.03 &  & ar-en & 27.1 & 47.0 & 76.54 \\
fr-nl & 10.8 & 31.0 & 58.50 &  & de-en & 27.8 & 44.4 & 77.57 \\
fr-ru & 8.6 & 26.7 & 57.07 &  & fr-en & 27.1 & 48.7 & 78.82 \\
fr-zh & 15.9 & 15.8 & 70.78 &  & nl-en & 22.6 & 42.2 & 74.25 \\
nl-ar & 2.2 & 15.4 & 63.47 &  & ru-en & 25.6 & 44.2 & 74.46 \\
nl-de & 10.2 & 28.5 & 53.65 &  & zh-en & 23.5 & 45.7 & 77.04 \\
nl-fr & 14.4 & 34.4 & 66.55 &  & avg. & 25.62 & 45.37 & 76.45 \\
nl-ru & 5.3 & 19.3 & 55.53 &  &  &  &  &  \\
nl-zh & 5.5 & 6.2 & 58.77 &  &  &  &  &  \\
ru-ar & 6.5 & 16.0 & 62.69 &  &  &  &  &  \\
ru-de & 6.1 & 24.3 & 52.89 &  &  &  &  &  \\
ru-fr & 18.2 & 39.0 & 69.95 &  &  &  &  &  \\
ru-nl & 6.3 & 22.5 & 54.36 &  &  &  &  &  \\
ru-zh & 7.6 & 13.3 & 61.94 &  &  &  &  &  \\
zh-ar & 8.7 & 26.5 & 70.88 &  &  &  &  &  \\
zh-de & 3.0 & 19.5 & 50.82 &  &  &  &  &  \\
zh-fr & 17.7 & 39.7 & 73.56 &  &  &  &  &  \\
zh-nl & 4.4 & 19.3 & 54.20 &  &  &  &  &  \\
zh-ru & 4.1 & 19.5 & 50.47 &  &  &  &  &  \\
\hline
avg. & 8.86 & 23.30 & 60.70 &  &  &  &  &  \\
seen$\rightarrow$seen & 14.77 & 27.80 & 71.07 &  & en$\rightarrow$seen & 15.80 & 28.47 & 72.35 \\
seen$\rightarrow$unseen & 6.31 & 23.08 & 54.81 &  & en$\rightarrow$unseen & 17.47 & 35.00 & 65.24 \\
unseen$\rightarrow$seen & 8.61 & 20.24 & 63.81 &  & seen$\rightarrow$en & 25.90 & 47.13 & 77.47 \\
unseen$\rightarrow$unseen & 7.15 & 23.70 & 54.51 &  & unseen$\rightarrow$en & 25.33 & 43.60 & 75.43 \\
\bottomrule
\end{tabular}
}
\caption{Detailed results of BLOOMZ-7b1 fine-tuned with MT+Align for BLOOMZ+3 de-nl-ru.}
\label{7b1-3langs-AD}
\end{table*}

\begin{table*}
\centering
\resizebox{\linewidth}{!}{
\begin{tabular}{lrrrc|lrrr}
\toprule
Zero-shot & BLEU & chrF++ & COMET &  & Supervised & BLEU & chrF++ & COMET \\
\toprule
ar-de & 6.9 & 24.7 & 58.10 &  & en-ar & 14.6 & 35.6 & 76.70 \\
ar-fr & 26.2 & 48.2 & 74.96 &  & en-de & 20.4 & 36.0 & 65.96 \\
ar-nl & 8.8 & 24.7 & 59.53 &  & en-fr & 27.9 & 50.0 & 77.65 \\
ar-ru & 6.5 & 22.7 & 55.33 &  & en-nl & 14.8 & 34.8 & 63.11 \\
ar-zh & 28.6 & 22.3 & 77.64 &  & en-ru & 13.3 & 29.0 & 61.43 \\
de-ar & 3.3 & 19.8 & 68.27 &  & en-zh & 36.1 & 27.7 & 80.31 \\
de-fr & 15.2 & 35.8 & 67.05 &  & avg. & 21.18 & 35.52 & 70.86 \\
de-nl & 8.2 & 26.0 & 53.35 &  &  &  &  &  \\
de-ru & 4.4 & 17.9 & 54.79 &  &  &  &  &  \\
de-zh & 12.0 & 9.9 & 65.20 &  &  &  &  &  \\
fr-ar & 14.2 & 35.2 & 74.84 &  &  & BLEU & chrF++ & COMET \\
fr-de & 8.9 & 28.4 & 53.81 &  & ar-en & 33.7 & 53.5 & 79.81 \\
fr-nl & 10.1 & 29.9 & 56.92 &  & de-en & 27.1 & 43.9 & 77.04 \\
fr-ru & 8.1 & 26.0 & 55.96 &  & fr-en & 29.6 & 51.0 & 79.60 \\
fr-zh & 30.2 & 25.6 & 79.43 &  & nl-en & 22.0 & 41.4 & 73.54 \\
nl-ar & 3.1 & 18.2 & 67.72 &  & ru-en & 25.1 & 43.9 & 74.05 \\
nl-de & 10.4 & 27.7 & 52.67 &  & zh-en & 32.6 & 54.3 & 79.75 \\
nl-fr & 16.9 & 37.3 & 68.46 &  & avg. & 28.35 & 48.00 & 77.30 \\
nl-ru & 4.8 & 17.8 & 54.71 &  &  &  &  &  \\
nl-zh & 8.1 & 7.0 & 63.96 &  &  &  &  &  \\
ru-ar & 11.9 & 31.5 & 72.45 &  &  &  &  &  \\
ru-de & 6.1 & 23.7 & 52.74 &  &  &  &  &  \\
ru-fr & 21.2 & 42.5 & 71.71 &  &  &  &  &  \\
ru-nl & 6.8 & 22.6 & 53.91 &  &  &  &  &  \\
ru-zh & 21.3 & 20.7 & 74.63 &  &  &  &  &  \\
zh-ar & 13.1 & 34.1 & 74.92 &  &  &  &  &  \\
zh-de & 4.1 & 22.3 & 52.13 &  &  &  &  &  \\
zh-fr & 23.8 & 46.5 & 76.54 &  &  &  &  &  \\
zh-nl & 4.8 & 19.9 & 54.26 &  &  &  &  &  \\
zh-ru & 5.7 & 21.9 & 50.60 &  &  &  &  &  \\
\hline
avg. & 11.79 & 26.36 & 63.22 &  &  &  &  &  \\
seen$\rightarrow$seen & 22.68 & 35.32 & 76.39 &  & en$\rightarrow$seen & 26.20 & 37.77 & 78.22 \\
seen$\rightarrow$unseen & 7.10 & 24.50 & 55.18 &  & en$\rightarrow$unseen & 16.17 & 33.27 & 63.50 \\
unseen$\rightarrow$seen & 12.56 & 24.74 & 68.83 &  & seen$\rightarrow$en & 31.97 & 52.93 & 79.72 \\
unseen$\rightarrow$unseen & 6.78 & 22.62 & 53.69 &  & unseen$\rightarrow$en & 24.73 & 43.07 & 74.88 \\
\bottomrule
\end{tabular}
}
\caption{Detailed results of BLOOMZ-7b1 fine-tuned with MTInstruct for BLOOMZ+3 ar-de-fr-nl-ru-zh.}
\label{7b1-6langs-A}
\end{table*}

\begin{table*}
\centering
\resizebox{\linewidth}{!}{
\begin{tabular}{lrrrc|lrrr}
\toprule
Zero-shot & BLEU & chrF++ & COMET &  & Supervised & BLEU & chrF++ & COMET \\
\toprule
ar-de & 6.7 & 24.2 & 57.45 &  & en-ar & 15.1 & 35.8 & 76.76 \\
ar-fr & 27.5 & 49.2 & 75.21 &  & en-de & 20.6 & 35.9 & 65.88 \\
ar-nl & 8.7 & 24.8 & 59.14 &  & en-fr & 27.5 & 49.4 & 77.46 \\
ar-ru & 6.7 & 21.6 & 55.04 &  & en-nl & 15.0 & 35.6 & 63.70 \\
ar-zh & 30.1 & 24.4 & 78.54 &  & en-ru & 13.5 & 29.5 & 61.62 \\
de-ar & 3.5 & 19.7 & 68.39 &  & en-zh & 36.3 & 27.7 & 80.52 \\
de-fr & 15.4 & 35.8 & 67.81 &  & avg. & 21.33 & 35.65 & 70.99 \\
de-nl & 9.6 & 27.3 & 53.74 &  &  &  &  &  \\
de-ru & 4.7 & 17.9 & 54.23 &  &  &  &  &  \\
de-zh & 12.0 & 9.9 & 65.40 &  &  &  &  &  \\
fr-ar & 14.9 & 36.3 & 74.98 &  &  & BLEU & chrF++ & COMET \\
fr-de & 9.2 & 28.3 & 52.96 &  & ar-en & 33.9 & 53.7 & 79.74 \\
fr-nl & 11.3 & 31.1 & 57.62 &  & de-en & 27.1 & 43.6 & 77.13 \\
fr-ru & 8.8 & 26.2 & 56.31 &  & fr-en & 29.7 & 51.0 & 80.03 \\
fr-zh & 31.1 & 26.9 & 79.93 &  & nl-en & 22.6 & 42.3 & 73.94 \\
nl-ar & 3.3 & 18.5 & 68.02 &  & ru-en & 25.8 & 44.5 & 74.07 \\
nl-de & 9.4 & 26.5 & 52.33 &  & zh-en & 32.5 & 54.5 & 80.01 \\
nl-fr & 17.2 & 37.3 & 68.38 &  & avg. & 28.60 & 48.27 & 77.49 \\
nl-ru & 4.4 & 17.1 & 53.63 &  &  &  &  &  \\
nl-zh & 8.3 & 7.0 & 64.08 &  &  &  &  &  \\
ru-ar & 12.4 & 32.1 & 72.40 &  &  &  &  &  \\
ru-de & 5.7 & 22.9 & 51.90 &  &  &  &  &  \\
ru-fr & 21.5 & 42.7 & 72.08 &  &  &  &  &  \\
ru-nl & 6.3 & 22.1 & 53.32 &  &  &  &  &  \\
ru-zh & 22.7 & 24.6 & 75.36 &  &  &  &  &  \\
zh-ar & 13.9 & 35.4 & 75.68 &  &  &  &  &  \\
zh-de & 3.6 & 21.3 & 51.32 &  &  &  &  &  \\
zh-fr & 24.5 & 47.0 & 76.98 &  &  &  &  &  \\
zh-nl & 4.9 & 20.3 & 54.30 &  &  &  &  &  \\
zh-ru & 5.5 & 21.1 & 50.49 &  &  &  &  &  \\
\hline
avg. & 12.13 & 26.65 & 63.23 &  &  &  &  &  \\
seen$\rightarrow$seen & 23.67 & 36.53 & 76.89 &  & en$\rightarrow$seen & 26.30 & 37.63 & 78.25 \\
seen$\rightarrow$unseen & 7.27 & 24.32 & 54.96 &  & en$\rightarrow$unseen & 16.37 & 33.67 & 63.73 \\
unseen$\rightarrow$seen & 12.92 & 25.29 & 69.10 &  & seen$\rightarrow$en & 32.03 & 53.07 & 79.93 \\
unseen$\rightarrow$unseen & 6.68 & 22.30 & 53.19 &  & unseen$\rightarrow$en & 25.17 & 43.47 & 75.05 \\
\bottomrule
\end{tabular}
}
\caption{Detailed results of BLOOMZ-7b1 fine-tuned with MT+Align for BLOOMZ+3 ar-de-fr-nl-ru-zh.}
\label{7b1-6langs-AD}
\end{table*}

\end{document}